\documentclass[11pt,oneside,reqno]{amsart}
\usepackage{algorithm,algorithmicx,algpseudocode}
\usepackage{epsfig,psfrag,subfigure}
\usepackage{epstopdf}
\usepackage{amsmath,amssymb,amsfonts,latexsym, mathrsfs,euscript}
\usepackage{graphicx,graphics,cite}
\usepackage{wrapfig}
\usepackage[usenames]{color}
\usepackage{geometry}   
\usepackage{subfig}             
\usepackage{colortbl,booktabs,ctable,multirow}
\usepackage{epstopdf}
\usepackage{subfig}

\usepackage[T1]{fontenc}

\DeclareCaptionLabelFormat{andtable}{#1~#2  \&  \tablename~\thetable}

\DeclareGraphicsExtensions{.pdf,.png,.jpg}

\topmargin by -\headsep \textheight 8.7in \oddsidemargin 0.1in
\newtheorem{theorem}{\bf Theorem}

\newcommand{\beq}{\begin{equation}}
\newcommand{\eeq}{\end{equation}}
\newcommand{\beqa}{\begin{eqnarray}}

\newcommand{\eeqa}{\end{eqnarray}}

\newcounter{nfig}

\newcommand{\ignore}[1]{}

\newcommand\ds{\displaystyle}

\newcommand\bs{\boldsymbol}

\newtheorem{note}{Note}

\theoremstyle{definition}
\newtheorem{definition}{Definition}

\numberwithin{equation}{section}

\title[pattern recognition and abnormality detection]{A Landmark-based algorithm for automatic pattern recognition and abnormality detection}

\author{S. Huzurbazar}
\address{Department of Statistics, University of Wyoming, Laramie, WY 82071-3036, USA}
\email{lata@uwyo.edu\\ TEL:307-766-4826}

\author{Dongyang Kuang}
\address{Department of Mathematics, University of Wyoming, Laramie, WY 82071-3036, USA}
\email{dykuang@outlook.com\\ TEL:307-766-4221}

\author{Long Lee}
\address{Department of Mathematics, University of Wyoming, Laramie, WY 82071-3036, USA}
\email{llee@uwyo.edu\\ TEL:307-766-4368}

\begin{document}
\maketitle

\begin{abstract}
We study a class of mathematical and statistical algorithms with the aim of establishing a computer-based framework for fast and reliable automatic abnormality detection on landmark represented image templates. Under this framework, we apply a landmark-based algorithm for finding a group average as an estimator that is said to best represent the common features of the group in study.  This algorithm extracts information of momentum at each landmark through the process of template matching. If ever converges, the proposed algorithm produces a local coordinate system for each member of the observing group, in terms of the residual momentum. We use a Bayesian approach on the collected residual momentum representations for making inference. For illustration, we apply this framework to a small database of brain images for detecting structure abnormality. The brain structure changes identified by our framework are highly consistent with studies in the literature.

\end{abstract}

\begin{description}
\item[{\footnotesize\bf keywords}]
{\footnotesize Pattern recognition, abnormality detection, group average, landmark, template matching, \\ momentum, posterior predictive analysis, brain images, structure changes.}
\end{description}


\section{Introduction}

In the past 60 years, image analysis and the related science have developed at an enormous speed along with computational power. Abnormality detection, an active field of research, can be very useful if key features can be extracted from the image templates under comparison. The process of identifying prominent features generally includes two major steps: (1) transforming the image or shape representations from nonlinear spaces into locally linear coordinates for any necessary calculations, and (2)  making inferences from these calculations via statistical analyses.  Among various representations used, landmark representation, despite its long history and simplicity, has proven to be one of most effective tools in making statistical inferences on the shape/image objects. 

The traditional Procrustes methods extract residual vectors of landmark locations through calculating a group average, which serves as a quantity similar to the statistical mean\cite{bib:Bookstein, bib:Bookstein_schizo}. The resulting residual vectors are then analyzed 
via classical statistical techniques for making inferences.
Alternatively, a class of diffeomorphic methods, that were first introduced in the 1990s \cite{bib:Grenander93, bib:dgm98} and developed, both theoretically and numerically more recently, \cite{bib:bmty05,  bib:jm00, bib:bk06, anatomy, bib:mty06, image2, bib:eqs, bib:MM2} model the given image data through certain manifolds and relate two deformable templates through geodesics on the selected manifolds. For detailed information about the development of this class of methods, we refer readers to the book and paper by Younes et al. \cite{bib:book_Younes, bib:survey}.
The geodesic of a Riemannian manifold is governed by a class of partial differential equations (PDE)  called the Euler-Poincar\'e (EP) equations \cite{bib:cdm12, bib:hrty04, bib:mumford, bib:eqs, NP}.  When images or shapes are represented by landmarks, geodesic matching algorithms can be constructed via finding the proper initial conditions of the equivalent finite dimensional particle system of the EP equations.  For example, an efficient  template matching algorithm that takes advantage of the landmark representation and the particle formulation of the EP equation was introduced by Camassa et al. \cite{bib:ckl14, bib:ckl15}.  This method uses a constant matrix to update the search direction of the geodesic shooting for finding the initial conditions that evolve the reference template to the target one, instead of the traditional methods of forward-backward integration for updating the gradient (known as the LDDMM) or Newton's optimization. Another feature of this method is that a non-smooth conic kernel for the particle system is used to accelerate the convergence of matching.
We give a brief introduction of the aforementioned geodesic matching algorithm in Section \ref{sec:basic}. Using this matching algorithm as our underlying warping method,  in this paper we propose : (1) a landmark-based algorithm calculating the residual momentum representation for each landmark template as feature vectors and (2) a Bayesian approach on the resulting feature vectors for detecting the most statistically significant landmarks that encode the difference between groups under comparison. While calculating the momentum representation, like in the Procrustes approach, an estimator as an average of a group on a general Riemannian manifold is also calculated as a byproduct. Some similar estimators of central tendency in this context are also given in \cite{bib:bk06, bib:fvj08}.  
The resulting residual momentum at convergence then provides a locally linear space for applying our Bayesian statistical analysis for detection purposes. This is the novelty of our proposed algorithm as well as the major difference that sets it apart from the existing literature \cite{image2, xavier, ma,joshi-ssa,bib:fvj08}.

The paper is organized as follows. In Sections \ref{sec:basic} and \ref{sec:group}, an averaging algorithm on the momentum field under a deformative template setting is constructed using a landmark representation and the particle shooting algorithm. We show how the momentum representation for each template is collected during the averaging procedure. 
In Section \ref{sec:detect}, we present an example of practical usage of the proposed algorithm for medical image abnormality detection.  Using the momentum representation based on the calculated average template, a Bayesian predictive statistical approach is presented to detect landmarks whose momentum coordinates encode the most significant differences between groups under comparison. We call this set of landmarks a predictor. Finally, we show that the brain structure changes identified by our predictors are highly consistent with studies in the literature.   


\section{Basic Formulation}\label{sec:basic}
For template matching, based on the principle of diffeomorphism, the warping between two images is established through the geodesic defined on the diffeomorphism group Diff$(\mathbb{R}^n)$ or Diff$(D)$ for some $D\subseteq \mathbb{R}^n$, which is an open subset of the space $I_d+C_c^\infty(\mathbb{R}^n,\mathbb{R}^n)$, where  $I_d$ is an identity map. Suppose that such a space is denoted by $\mathcal{G}$.  
For ${\bs{u}} \in T_e\mathcal{G}$, a vector in the tangent space of $\mathcal{G}$ with identity $e\in\mathcal{G}$, its norm is defined by:
\begin{equation}
\Vert {\bs{u}} \Vert_{\scriptscriptstyle{L}}^2:=<{\bs{u},\bs{u}}>_{\scriptscriptstyle{L}}\, :=\, \int_{D}L{\bs u}\cdot{\bs u}\,\,d{\bs x}.
\end{equation}
Suppose that the operator $L$ is defined by $L=(I_d-a^2\Delta)^b,\,b \geq 1$, the associated Green's kernel of $L$ is given by
\begin{equation}
G_{b-n/2}(|\bs{x}|) = \frac{2^{n/2- b}}{(2\pi a)^{n/2}a^{b}\Gamma(b)}|\bs{x}|^{b -n/2}K_{b-n/2}\left(\frac{|\bs{x}|}{a}\right),
\end{equation}
where, $K_{b -n/2}$ is the modified Bessel function of the second kind of order $b -n/2$ and $\Gamma(b)$ is the usual notation for the Gamma function~\cite{bib:mumford}.  A notable special parametric choice is the two-dimensional ($n=2$) Green function's with $a=1$ and $ b=3/2$, resulting in the simple form\footnote{In this case the space is $I_d+C_0^0(\mathbb{R}^n,\mathbb{R}^n)$} 
\beq\label{eq:nu=3/2}
G_{1/2}(|\bs{x}|) = \frac{1}{2\pi}e^{-|\bs{x}|} \, . 
\eeq
This yields a conic shape kernel that has advantages for template matching as discussed in \cite{bib:ckl15};  we will use this later in our numerical experiments.  

The length of a curve $\phi: [0,1] \rightarrow \mathcal{G}$ is defined by using the right translation 
\begin{equation}
l(\phi)=\int_0^1\Vert {\bs{u}}\Vert_{\scriptscriptstyle{L}} dt, \quad \text{for} \, \bs{u}=\frac{\partial \phi}{\partial t}({\phi^{-1}({\bs{x}},t),t}).
\end{equation}
Furthermore, an energy functional can be defined by
\begin{equation}
E(\phi)=\int_0^1<{\bs{u}},{\bs{u}}>_{\scriptscriptstyle{L}}dt.
\end{equation}
The geodesic is usually obtained by minimizing the energy functional $E$ rather than the length $l$. This is because minimizing $E$ chooses not only the best path but also the simplest parametrization of time in $\phi$ that satisfies the constraints $\phi({\bf{x}},0)=I_0$ and $\phi({\bf{x}},1)=I_1$, where $I_0$ is referred to as the reference,  whereas $I_1$ is the target \cite{bib:mumford}. In this case, we have $l(\phi) =E(\tilde{\phi})= l(\tilde{\phi})^2$, where $\tilde{\phi}=\phi(\bs x, f(t))$ with $f$ being an invertible increasing function (a reparametrization of time) from [0,1] to itself.  Note that we can assume the existence and the uniqueness of such a geodesic in our problems, as long as the manifold we consider is compact and we only work on some neighborhood of $e \in \mathcal{G}$. 

\begin{figure}[tbh]
\includegraphics[scale=.5]{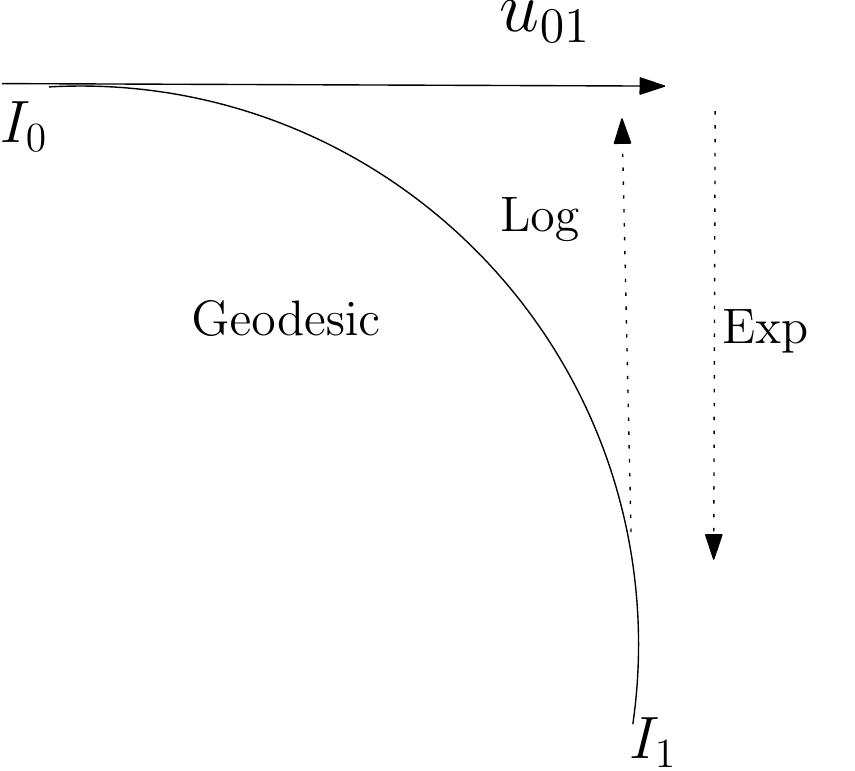}
\caption{A schematic for Riemann maps between the manifold and the tangent space. $\bs u_{01}$ serves as the "shape coordinate" for template $I_1$ based on $I_0$. Through this map, the information on the nonlinear manifold is extracted and mapped to the tangent space that is tangent to the manifold at $I_0$. The tangent space is linear, on which statistical analysis can easily be applied.}
\label{fig:explog}
\end{figure}

The Euler-Lagrange equation, arising from minimizing the energy functional, turns out to have exactly the same form as the EP equations \cite{anatomy,bib:mumford}:
\begin{equation}\label{EPDiff}
{\bs m}_t + (\bs{u}\cdot\nabla)\bs{m}  +(\nabla\cdot\bs{u})\bs{m}  + (\nabla\bs{u}^{t})\bs{m}= 0,
\end{equation}
with ${\bs m}=L{\bs u}$, subject to the ``time'' boundary conditions $\phi({\bf{x}},0)=I_0$ and $\phi({\bf{x}},1)=I_1$. Assuming that the space is geodesic complete, we can convert this boundary-value problem into finding the corresponding initial condition for an initial-value problem.
  Let the corresponding initial condition $\bs u_{01}$ be defined through a proper logarithmic map from M to $T_{I_0}M$, the tangent space of $M$ at $I_0$, by:
\begin{equation}
\bs u_{01} = \text{Log}_{I_0} (I_1).
\end{equation} 
The corresponding exponential map from $T_{I_0}M$ to M then can be defined accordingly by
\begin{equation}
I_1 = \text{Exp}_{I_0} (\bs u_{01}).
\end{equation}
The Exp map is an isometry between $T_{I_0}M$ and $M$ to the first order \cite{bib:mumford}, here we take the results from finite-dimensional Riemannian geometry since this paper will only deal with finite landmarks:
\begin{equation}\label{eqn:Mtaylor}
d(I_i,I_j)^2 = ||\bs u_{0i}-\bs u_{0j}||^2-\frac{1}{3} ||\bs u_{0i} \wedge \bs u_{0j} ||^2_R +o(\text{max}(||\bs u_{0i}||,||\bs u_{0j}||)^4),
\end{equation}
where $||\cdot ||_R$ is the norm of curvature tensor and $I_i,I_j \in M$. By letting $i=0, j=1$, $\bs u_{00} = \bs 0$, the term $\frac{1}{3} ||\bs u_{0i} \wedge \bs u_{0j} ||^2_R = 0$. Combining all the previous results, we will have 
\begin{equation}\label{eq:l}
l(\phi) =E(\tilde{\phi})= l(\tilde{\phi})^2 = d(I_0,I_1)^2 = ||\bs u_{01}||^2 + o(||\bs u_{01}||^4).
\end{equation} 
Thus, numerically, the length of tangent vector can be used to estimate the length of the geodesic. This first-order isometry is illustrated in Figure \ref{fig:explog}.

Under the setting of landmark representation, a diffeomorphism $\phi$ can be represented by its actions on the landmark space $L^N$, i.e.
\beq
\phi: L^N \rightarrow L^N, \quad I:=\{{\bf{x}_i}\}_{i=1}^N\,\rightarrow \,\phi(I):=\{\phi({\bf{x}_i})\}_{i=1}^N.
\eeq
This gives a finite dimensional approximation of the original infinite functional space\footnote{Readers are referred to \cite{bib:unlabel2, bib:unlabel1} for methods dealing with non-labeled landmarks.}. 
In the landmark space $L^N$, it is convenient to define a pair of variables $(\bs q_i,\bs p_i)$ for each landmark $\bs x_i = \bs q_i$ and the corresponding momentum ${\bs p}_i$. For given initial data $\{(\bs q_i,\bs p_i)\}_{i=1}^N$, the time evolution of the Euler-Lagrangian (or the EP) equations is obtained by evolving the particle system in time for $(\bs q_i,\bs p_i)$ \cite{bib:ckl14}:
\beq
\begin{split}
\ds\frac{d \bs{q}_i}{dt}& =\sum \limits_{j=1}^N G_{ b-1}(|\bs{q}_i-\bs{q}_j|)\bs{p}_j,\\
\ds\frac{d\bs{p}_i}{dt}& = -\sum_{\substack{j=1\\ j\ne i}}^N(\bs{p}_i\cdot\bs{p}_j)G_{ b-1}'(|\bs{q}_i-\bs{q}_j|)\frac{\bs{q}_i-\bs{q}_j}{|\bs{q}_i -\bs{q}_j|}.
\end{split}\label{eq:particle}
\eeq
The above particle system gives an easy way of calculating the tangent vector, which is the velocity field ${\bs u}(\bs{x}, t)$ of the EP equations,  
\beq\label{eq:u}
\bs{u}(\bs{x}, t) = \sum_{j=1}^{N}G_{ b-1}(|\bs{x} - \bs{q}_j|)\bs{p}_j.
\eeq 
It is worth noting that numerically solving the particle system (\ref{eq:particle}) is equivalent to approximating an Exp map. Moreover, with this particle setting, on the tangent space, numerically it is advantageous for us to directly work on the momentum field $\bs p$, other than the velocity ${\bs u}$ for the Log map. Detailed discussion about the advantage of using the momentum field is in Section \ref{sec:group}.  Note that the conversion between the velocity and the momentum fields is straightforward by using Eq. (\ref{eq:u}).

\begin{figure}[tbh]
 \includegraphics[scale=.8]{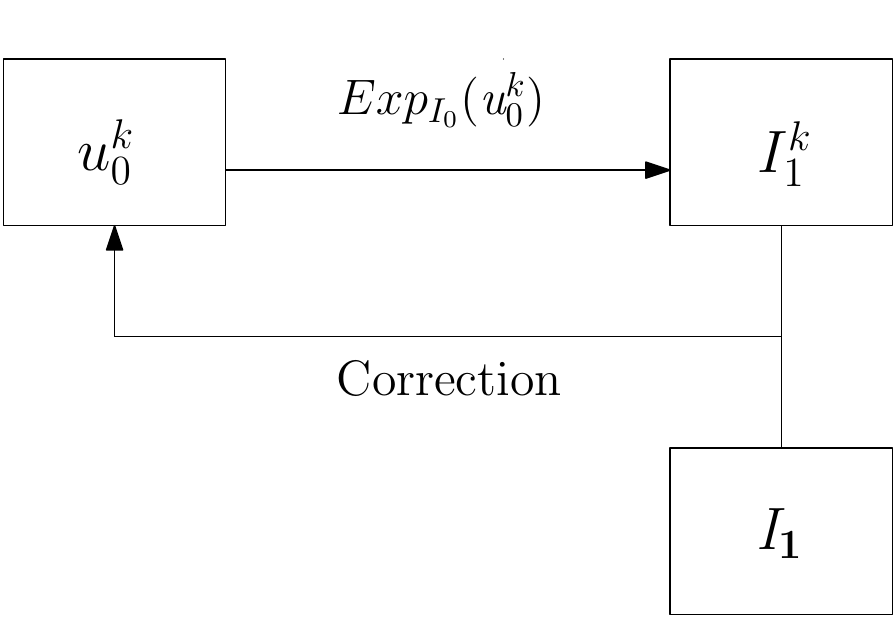}
 \caption{A schematic illustration for the shooting algorithm. In the $k^{\text{th}}$ iteration before convergence, the particle system is used to calculate the other end of the geodesic by adopting $(I_0, \bs u_0^k)$ as the initial data. A miss-fit function between $I_1^k$ and the target $I_1$ is calculated to adjust $\bs u_0^k$ for the next approximation $\bs u_0^{k+1}$.}
 \label{shoot}
 \end{figure}

For our applications, landmark positions $\{\bs x_i=\bs q_i\}_{i=1}^N$ at $t=0$ and $t=1$ are known, while the initial momenta $\{\bs p_i\}_{i=1}^N$ at $t=0$ are not. A  shooting algorithm that searches for the initial momenta is introduced in \cite{bib:ckl15}. In that algorithm, a prediction-correction step:
\begin{eqnarray}
\bs{u}^{k+1}_0&=&\bs{u}^k_0+h^k(\bs{I}^1-\bs{I}^k_1),\\
\bs{p}^{k+1}_0&=&\text{Convert}(\bs{u}^{k+1}_0; \bs{I}_0),
\end{eqnarray}
 where 
 \begin{align}
 \bs{u}^k_0&: \text{the $k^{\text{th}}$ iteration of the velocity field at}\, t=0 \,\,  \text{for} \,\, I_0,\\
 \bs{I}_1^k&: \text{the $k^{\text{th}}$ iteration of $I_1$ obtained by solving the particle system},\\
 h&: \text{the correction parameter at $k^{\text{th}}$ iteration,}
 \end{align}
is coupled with the particle system (\ref{eq:particle}) to search for the appropriate initial conditions $\bs p_i|_{t=0}$, $i=1,\cdots, N$, so that $\text{Exp}(\bs{I}_0;{\bs p}_i|_{t=0})=\bs{I}_1$. This shooting algorithm numerically generates the Riemann Log map. We refer readers to \cite{bib:ckl15} for more details about this algorithm. Figure \ref{shoot} gives a schematic illustration of this shooting algorithm.

\section{A geodesic shooting algorithm for an average}\label{sec:group}
In the literature, there are many classical approaches for obtaining an average of a group \cite{bib:Bookstein, bib:Bookstein_schizo, bib:mixed_ave, bib:mumford,joshi-ssa}.  The basic idea of these approaches is to calculate the Karcher mean or Fr\'echet mean with different choice of metrics \cite{bib:Karcher,bib:Fletcher}.  Recently more sophisticated estimators related to tangent spaces of Riemannian manifolds have been studied for their potential to be used as averages of groups \cite{bib:fvj08,image2,ini,xavier}. 
Among all these approaches, the classical morphormetric average, like the Procrustes average, in a sense, ignores the nonlinear and non-flat geometric features. On the other hand, to consider such features in finding an average under the setting of diffeomorphic deformation, it is necessary to find a nice linear space (at least locally), on which performing linear combinations is both mathematically sound and physically meaningful. To this end, we choose the momentum field as such a space for our proposed algorithm (described below), because the momentum provides a local template-based coordinate system which is also linear in nature \cite{bib:hrty04, xavier, image2}. 
As a result, in our algorithm, when we find an average of a group in the momentum field, and map the result through the Exp map back to the manifold, in principle certain information of the non-flat geometry will still be preserved.
At convergence, the initial momentum that drives the average to each individual observation will serve as the momentum representation of that landmark template and used for later statistical analysis.

\begin{figure}[tbh]
\includegraphics[scale=.6]{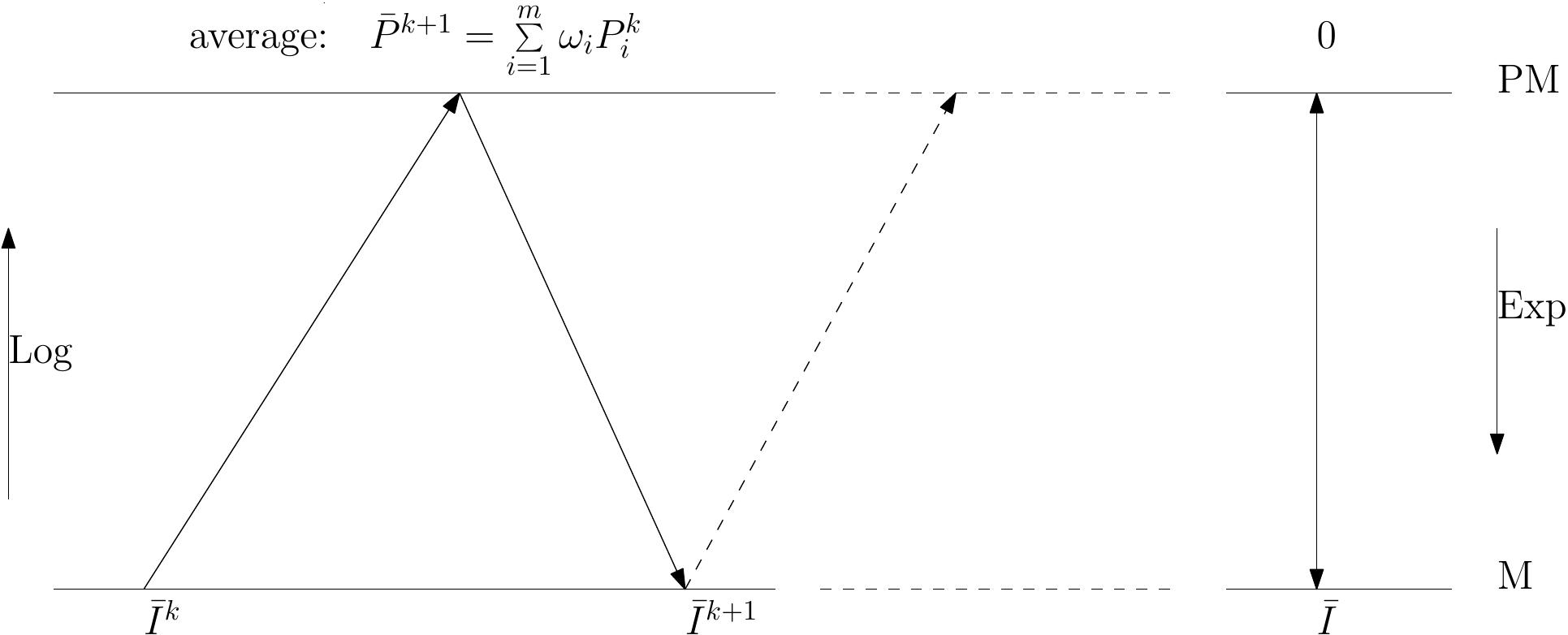}
\caption{The averaging scheme in brief. M stands for the manifold, PM denotes the momentum field, which is conjugate to TM (tangent field). As mentioned in the previous section, in the landmark setting, the Log mapping is calculated using a prediction-correction loop, while the Exp mapping can be done by solving the particle system \cite{bib:ckl15}. Once convergence is reached, $P_i$ will be used as residual momentum representation for each individual observation in the group.}
\label{averagescheme}
\end{figure}

Let $\overline{I}$ denote the average of the group $T:=\{I_k\}_{k=1}^m$.   We use Figure \ref{averagescheme} to illustrate the proposed averaging procedure, with details given in Algorithm \ref{alg:average}.

\begin{algorithm}[h]
\begin{algorithmic}

\State Choose an initial guess $\overline{I}^0 \notin T$ for the average;
\For{$k=0, 1, 2, \cdots$ until convergence}
\For{$i=0,1,2,\cdots,m$}

 \quad a) Extract tangent vector through Log mapping:
\begin{equation}
{\bs{u}}_i^{k} = \text{Log}_{\overline{I}^k}(I_i)
\end{equation} 

\quad b) Convert each tangent vector to momentum:
\begin{equation}
{\bs{p}}_i^k = \text{Convert} ({\bs{u}}_i^{k})
\end{equation}
\EndFor
\State Take the weighted average on the momentum field via:
\begin{equation}\label{eq:p_ave}
\overline{\bs{p}}^{k+1} = \sum\limits_{i=1}^m \omega_i{\bs{p}}_i^{k}
\end{equation}
\If{ $||\overline{\bs{p}}^{k+1} - \overline{\bs{p}}^{k}|| > \epsilon$}
\State Compute the next guess:
\begin{align}
\overline{\bs{u}}^{k+1} &= \text{Convert} (\overline{\bs{p}}^{k+1})\\
\overline{I}^{k+1} &= \text{Exp}_{\overline{I}^k}(\overline{\bs{u}}^{k+1})
\end{align}
\Else
  \State {\bf Break}; 
  \State {\bf Return} $\overline{I}^k$ as average $\overline{I}$.
\EndIf

\EndFor
\end{algorithmic}
\caption{Algorithm for finding an average of a group} 
\label{alg:average}
\end{algorithm}

 Note that in Algorithm \ref{alg:average}, $w_j$ in Eq. (\ref{eq:p_ave}) is called the $j^{\text{th}}$ {\em weight}. Two  special weighting schemes are worth noting. In the first scheme, the weight is chosen to be 
\beq\label{eq:w1}
w_i=1/m\quad \text{for all $i$'s.}
\eeq
This weighting scheme mimics the direct average method for the Karcher mean in flat Eulerian spaces, which gives the solution of the least {\em sum of distance squares}, i.e.
\begin{equation}\label{eqn:mind2}
\sum\limits_{i=1}^m d(\overline{I},I_i)^2.
\end{equation}
Note that for this weight, each member has an equal opportunity for contributing to the group average, which makes this scheme equivalent to that introduced in \cite{xavier}; the latter reference includes a thorough discussion of the so-called "center of mass" in Riemannian manifolds under a statistical setting.


The second weighting scheme is motivated by the idea of finding the {\em geometric median} \cite{bib:fvj08}, a robust estimator when the dataset has corrupted members. Using the idea from the Weizfeld's algorithm for Eulerian spaces \cite{bib:weizfeld}, the assigned weight for this scheme is calculated by 
\begin{equation}\label{eq:w2}
w_i = \frac{1}{d_i}/\sum\limits_{i=1}^m\frac{1}{d_i}.
\end{equation}
For this choice of weight, the estimator is the value that minimizes the {\em sum of distances}, denoted by 
\begin{equation} \label{eqn:mind}
D(\overline{I},T):=\sum\limits_{i=1}^m d(\phi_i),
\end{equation} 
where $\phi_i$ is the geodesic connecting the unknown group average $\overline{I}$ and each group member $I_i$, i.e. the restriction is now $\phi_i({\bf{x}},0)=\overline{I}$ and $\phi_i({\bf{x}},1)=I_i$ for each $i$.


The existence and uniqueness of both ``mean" ($\omega_i=\frac{1}{m}$) and ``median" ($w_i = \frac{1}{d_i}/\sum\limits_{i=1}^m\frac{1}{d_i}$) were investigated in \cite{bib:fvj08, xavier} respectively with assumptions on the manifold that either one of the below holds:
\begin{itemize}
	\item[(a)]the sectional curvature $\kappa<0$;
	\item[(b)]$0<\kappa<K$, and diam(U)<$\frac{\pi}{2\sqrt{K}}$ with $U$ a subset of the manifold that can cover all observations.
\end{itemize}
 Following the idea of Kuhn and others \cite{bib:weizfeld, bib:fvj08}, we now give a proof that shows the convergence of  Algorithm \ref{alg:average} for both choices of weights, (\ref{eq:w1}) and (\ref{eq:w2}) under the condition (b).
  
 \begin{theorem}[Local Convergence of  Algorithm \ref{alg:average}]\hfill \break
 \begin{enumerate}
  \item Algorithm \ref{alg:average} converges to the minimizer of (\ref{eqn:mind}) for $\omega_i = \frac{1}{||\bs u_i^k||_L}/\sum\limits_{i=1}^m \frac{1}{||\bs u_i^k||_L}$.
  \item Algorithm \ref{alg:average} converges to the minimizer of (\ref{eqn:mind2}) for $\omega_i = \frac{1}{m}$. 
 \end{enumerate}
 \end{theorem}
 \noindent
 {\em Proof:} 
 
If the condition (b) is satisfied, there exists a unique minimizer on the manifold \cite{bib:fvj08, xavier}. We only need to show the introduced algorithm actually converges under this condition.
 
We first show claim (1). Let $f(\bs x) = \sum\limits_{i=1}^m  d(\bs x ,\bs x_i )$, where $\bs x$ and $\{\bs x_i\}_{i=1}^m \in M$ and $M$ is a manifold. Let $\bs x,\bs y \in M$. The distance  $d(\bs x, \bs y)$ is defined by $||\text{Log}_{\bs x}(\bs y)||_{L,\bs x} =||\text{Log}_{\bs y}(\bs x)||_{L,\bs x}$, with $||\cdot||_{L,\bs x}$ the norm induced by operator $L$.
 
Since $f$ is a geodesic convex function, there will be no local optimum on $U$. In order to show that the algorithm converges, we only need to prove  $f(\hat{\bs x}^{k+1}) < f(\hat{\bs x}^k)$ if $f(\hat{\bs x}^k)$ is not the minimum.
 
 Let $\hat{\bs x}^{k+1}$ be the $k+1$ iteration of Algorithm \ref{alg:average}, with the following defined as: 
 \begin{equation}
 \bs u_i^{k+1}=\text{Log}_{\hat{\bs x}^{k+1}}(\bs x_i), \,\, \hat{\bs x}^{k+1} = \text{Exp}_{\hat{\bs x}^k}(\bs u^{k+1})\,\, \text{and} \,\,{\bs x}_i = \text{Exp}_{\hat{\bs x}^k}(\bs u_i^k).
 \end{equation}
Hence,
 \begin{align}
 f(\hat{\bs x}^{k+1})  
 = &\sum\limits_{i=1}^m d({\bs x}^{k+1},{\bs x}_i)\\
 = &\sum\limits_{i=1}^m d \left(\text{Exp}_{\hat{\bs {x}}^k} (\bs u^{k+1}) , \text{Exp}_{\hat{{\bs x}}^k} (\bs u_i^{k})\right)\\
 \leq & \sum\limits_{i=1}^m \Vert\bs u^{k+1}- \bs u_i^{k} \Vert_{L,\hat{\bs x}^{k}} . \label{eq:ieq1}
 \end{align}
 It is worth noting that the last inequality holds, only if all observations $x_i \in U$ and the condition (b) is satisfied. In this case, we can use Toponogov Comparison Theorem to obtain the inequality.

\begin{figure}[tbh]
 	\includegraphics[scale=0.7]{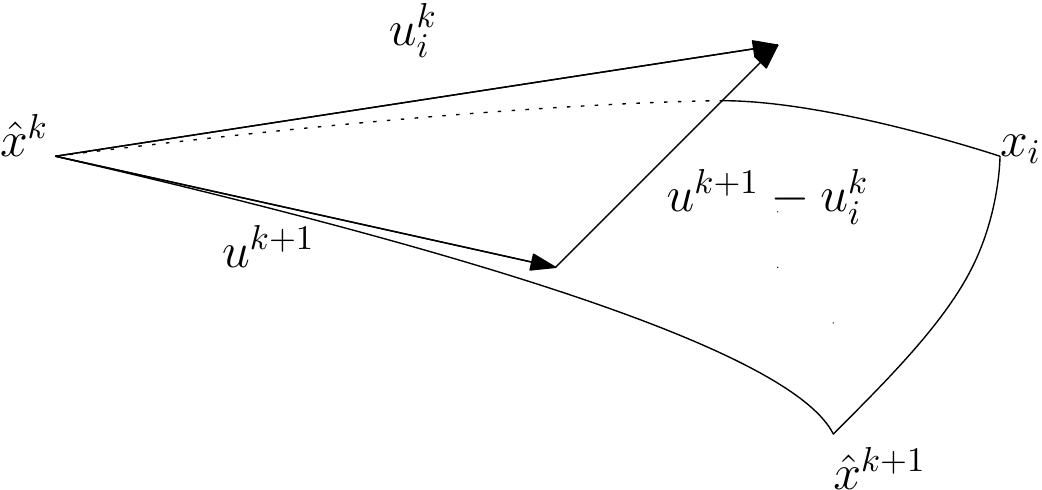}
 	\caption{Triangles on M and $T_{\hat{x}^k}$M}
 	\label{fig:trig}
 \end{figure}

Let $\Delta_{\hat{\bs x}^{k}\hat{\bs x}^{k+1}\bs x_i}$ be the geodesic triangle on the manifold, see Figure \ref{fig:trig}. Toponogov Comparison Theorem states that if observations ${\bs x}_i$ satisfy the condition (b), for each corresponding side of the two triangles, the distance on the $M$ is less than the corresponding distance on $T_{\hat{\bs x}^k}M$, i.e. $d( \hat{\bs x}^{k+1} , \bs x_i )\leq \Vert\bs u^{k+1}- \bs u_i^{k} \Vert_{L,\hat{\bs x}^{k}}$. 
 
Hereafter, for a fixed $\hat{\bs x}^k$, we drop $\hat{\bs x}^{k}$ in $||\cdot||_{L,\hat{\bs x}^{k}}$ and denote the norm by $||\cdot||_{L}$ for simplicity.  Suppose that the algorithm uses the update $\bs u^{k+1}=\sum\limits_{i=1}^m \omega_i \bs u_i^k$, where $\omega_i = \frac{1}{||\bs u_i^k||_L}/\sum\limits_{i=1}^m \frac{1}{||\bs u_i^k||_L}$, then $\bs u^{k+1}$ is the center of gravity of weights $\frac{1}{||\bs u_i^k||_L}$ placed at $\bs u_i^k$, and hence $\bs u^{k+1}$  minimizes the function\footnote{Since $||\bs u_i^k||_L$ is the denominator, observations ${\bs x}_i$ cannot be selected as an initial guess for starting the algorithm.}:
 \begin{equation}
 g(\bs u)=\sum\limits_{i=1}^m \frac{1}{||\bs u_i^k||_L} ||\bs u - \bs u_i^k||_L^2.
 \end{equation}
 Thus $g(\bs u^{k+1})\leq g(\bs 0)=\sum\limits_{i=1}^m ||\bs u_i^k||_L = f(\hat{\bs x}^k)$.
 
 On the other hand, 
 \begin{align}
 g(\bs u^{k+1})= &\sum\limits_{i=1}^m \frac{1}{||\bs u_i^k||_L} \big[\,(||\bs u^{k+1} - \bs u_i^k||_L-||\bs u_i^k||_L) +||\bs u_i^k||_L\,\big]^2\\
 = &  \sum\limits_{i=1}^m \frac{1}{||\bs u_i^k||_L} (||\bs u^{k+1} - \bs u_i^k||_L-||\bs u_i^k||_L)^2 + 2\sum\limits_{i=1}^m \Vert\bs u^{k+1}- \bs u_i^{k} \Vert_L - f(\hat{\bs x}^k).
 \end{align}
 Substituting this into $g(\bs u^{k+1})\leq f(\hat{\bs x}^k)$, we get
 \begin{align}
 &\sum\limits_{i=1}^m \frac{1}{||\bs u_i^k||_L} (||\bs u^{k+1} - \bs u_i^k||_L-||\bs u_i^k||_L)^2 + 2\sum\limits_{i=1}^m \Vert\bs u^{k+1}- \bs u_i^{k} \Vert_L \leq 2f(\hat{\bs x}^k).
 \end{align}
Hence, we have 
\beq\label{eq:ieq2} 
\sum\limits_{i=1}^m \Vert\bs u^{k+1}- \bs u_i^{k} \Vert_L < f(\hat{\bs x}^k).
\eeq
Now combining Eqs.  (\ref{eq:ieq1}) and (\ref{eq:ieq2}) yields
 \begin{equation}\label{eq:ieq3}
 f(\hat{\bs x}^{k+1}) \leq \sum\limits_{i=1}^m \Vert\bs u^{k+1}- \bs u_i^{k} \Vert_L < f(\hat{\bs x}^k).
 \end{equation}
Note that the equalities in (\ref{eq:ieq3}) hold if and only if  $\bs u^{k+1}=\bs 0$, which is equivalent to $\hat{\bs x}^{k+1}=\hat{\bs x}^k = \overline{\bs x}$, i.e. the iteration stops and the minimum is reached. This proves that Algorithm \ref{alg:average} converges when the weight (\ref{eq:w2}) is used. Moreover, at convergence, the result of Algorithm \ref{alg:average} minimizes the sum of the distances.
 
Following the above structure, and using the same notation, we can prove claim (2). Below, we repeat the proof but skip some steps to avoid redundancy. Redefining $f(\hat{\bs x})= \sum\limits_{i=1}^m d( \bs x , \bs x_i )^2$, we have  $f(\hat{\bs x}^{k+1}) \leq \sum\limits_{i=1}^m \Vert\bs u^{k+1}- \bs u_i^{k} \Vert_L^2 $. Since $\omega_i = \frac{1}{m}$, the function $g$ is now defined by \\ $g(\bs u)=\sum\limits_{i=1}^m ||\bs u - \bs u_i^k||_L^2 = f(\bs u)$. So by the same argument, we have
 $$f(\hat{\bs x}^{k+1}) \leq g(\hat{\bs x}^{k+1}) \leq g(\bs 0) = \sum\limits_{i=1}^m ||\bs u_i^k||_L^2 = f(\hat{\bs x}^k).$$
The equality holds only if $\hat{\bs x}^{k+1}=\hat{\bs x}^k$. This proves the convergence of Algorithm \ref{alg:average} with equal weights, and the result gives the least sum of distances squared. 
 $\square$\\

We remark that similar to Weizfeld's algorithm for flat Eulerian spaces, the initial guess for Algorithm \ref{alg:average} cannot be chosen as any of the group members $I_i$, otherwise the algorithm will not converge. However, there is a way to get around with this restriction in flat Eulerian spaces \cite{ini}, so that the algorithm converges for any initial guess. The extension of it on general Riemannian manifolds is still under our investigation. We also remark that Pennec \cite{xavier} investigated other geometric measurements in a statistical setting for the equally weighted schemes, see \cite{xavier} for further details. 

The two weighting schemes give rise to different average templates for a group, based on which, the resulting residual momentum representation for each group member will be different. We remark that the result of Theorem \ref{alg:average} relies heavily on the condition (b) which may not hold for every dataset, as discussed in \cite{bib:sec_curv}. Nevertheless, in this paper, we intend to use the algorithm as a feature extraction tool, a pre-processor to extract momentum features for the purpose of detecting local changes that are statistically significant. The data we use are often recorded from the same or similar objects, such as the annotated brain landmarks of different people or of the same people but obtained at different times. In these cases, observations are assumed ``close", and hence convergence can be expected.
In other cases, the algorithms may not converge\footnote{For example, curvatures are negative.}, or the minimizer is not unique.\footnote{For example, curvatures are positive, but diameter inequality is not satisfied for observations.} In such cases, we will use a smaller scale parameter $a$ in Green's kernel\footnote{All the following examples are conducted using $a^2=1$ unless specifically specified otherwise.} $\frac{1}{2\pi}e^{\frac{|\bs{x}|}{a^2}}$, so that the matrix $\{G_{ij}\}$ with entry $G_{ij}=\frac{1}{2\pi}e^{\frac{|\bs{x_i-x_j}|}{a^2}}$ will become more diagonally dominated when used for calculating the geodesic. Landmarks in this case will become less correlated with each other and the deformation will look more linear. By doing so, the geometry of the landmark deformation may become closer to flat case, and we can hope that the diameter inequality in condition (b) could then be satisfied and the proposed algorithms can at least converge  numerically.



\subsection{Example}
 
\begin{figure}[tbh]
\includegraphics[scale=.5]{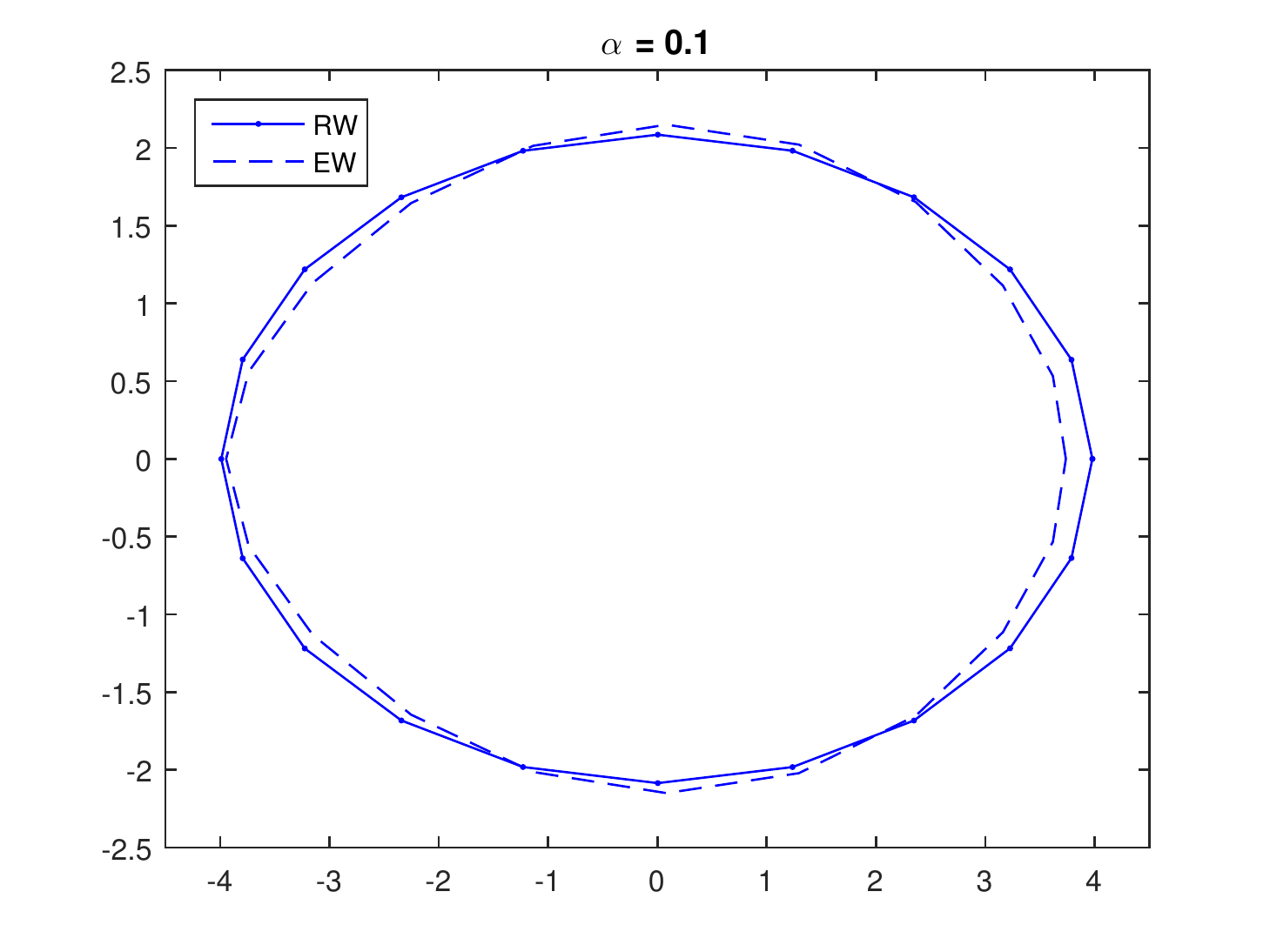}
\includegraphics[scale=.5]{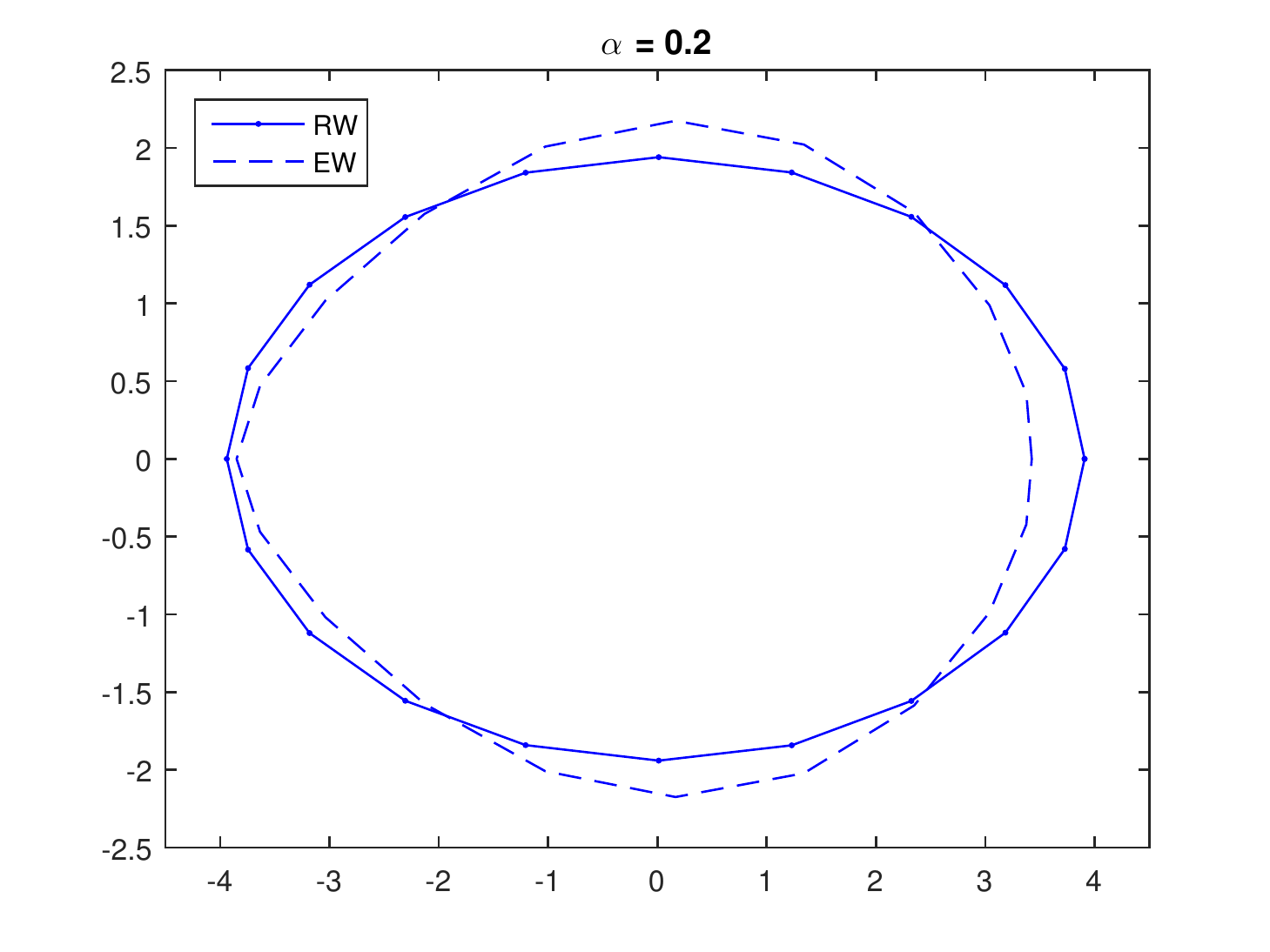}
\includegraphics[scale=.5]{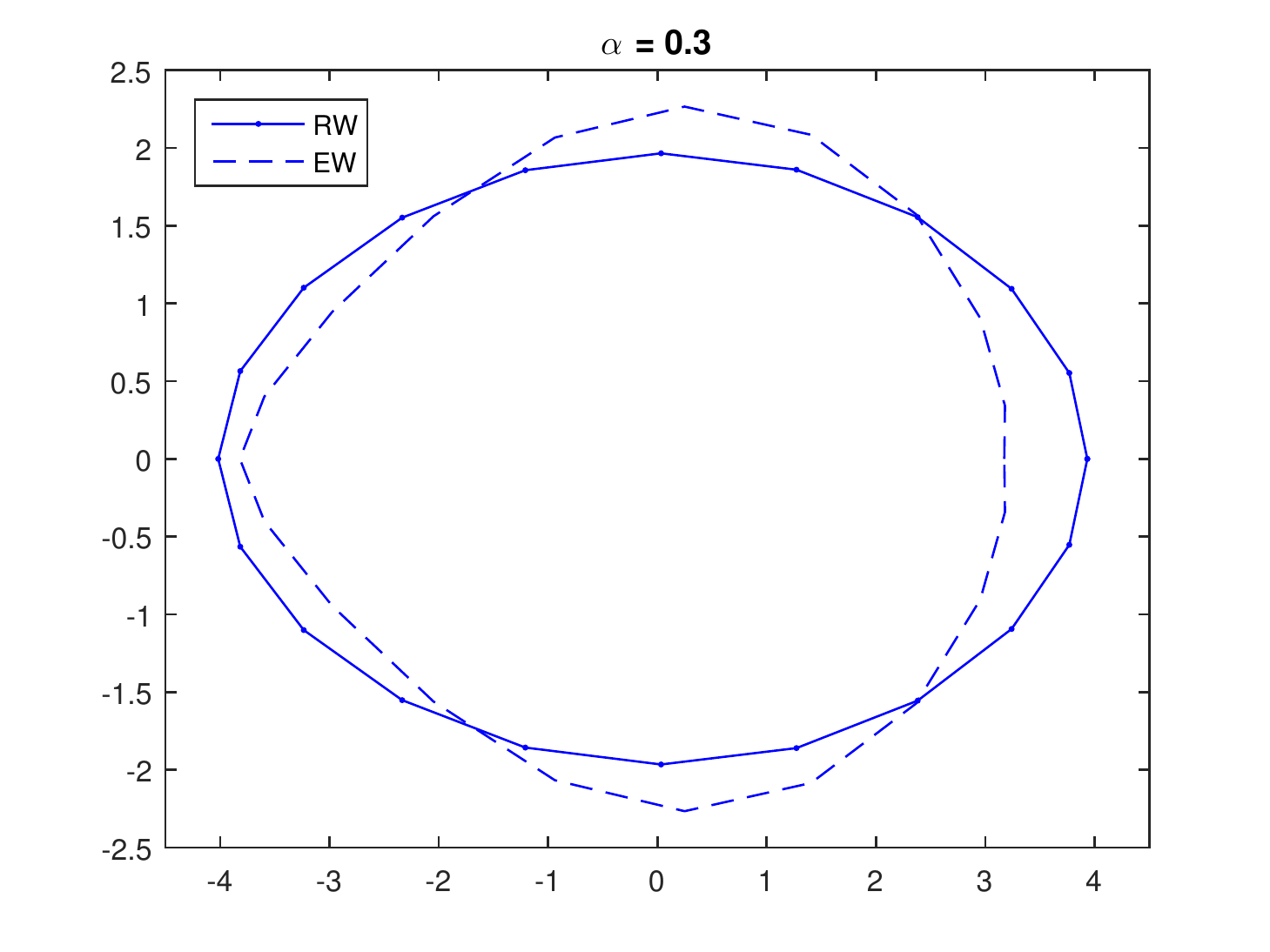}
\includegraphics[scale=.5]{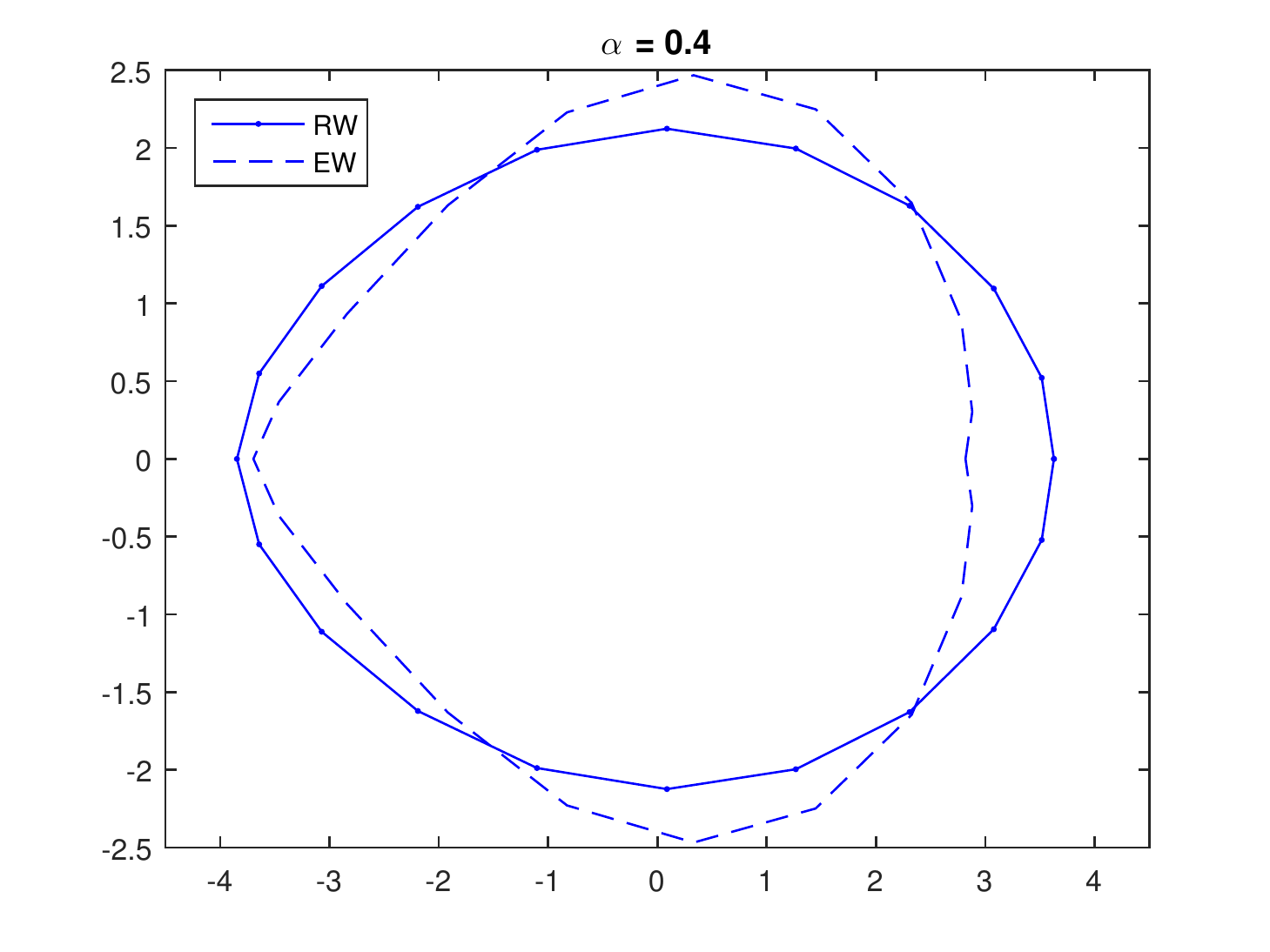}
\caption{Comparison of results from the two weighting schemes. "RW" stands for the robust weighting scheme, while "EW" is short for equally weighted scheme. The EW scheme are more sensitive to the fluctuations of the group components.}\label{fig:REW}
\end{figure}

In this example, we calculate the average for a group of 20 shapes by using the two different weighting schemes, respectively. 20 landmarks are used to represent each shape. In the group we consider, a portion of the group ($1-\alpha$, where $0 < \alpha < 1$), the shapes are the ellipses $\frac{x^2}{a^2}+\frac{y^2}{b^2}=1$, where $(a,b)$ obeys a normal distribution: $\mathcal{N}\big((4,2),0.2I\big)$, where $I$ is the identity matrix, whereas the rest of the group are the so-called outliers, which are copies of the heart-shape curve: 
\beq \label{eq:heart}
  \begin{cases}
 x =&\frac{1}{5}(13\cos(\theta)-5\cos(2\theta)-2\cos(3\theta)-\cos(4\theta))\\
 y =&\frac{1}{5}(16\sin(\theta)^3).
 \end{cases}
\eeq 
\begin{figure}[tbh]
\includegraphics[scale=.5]{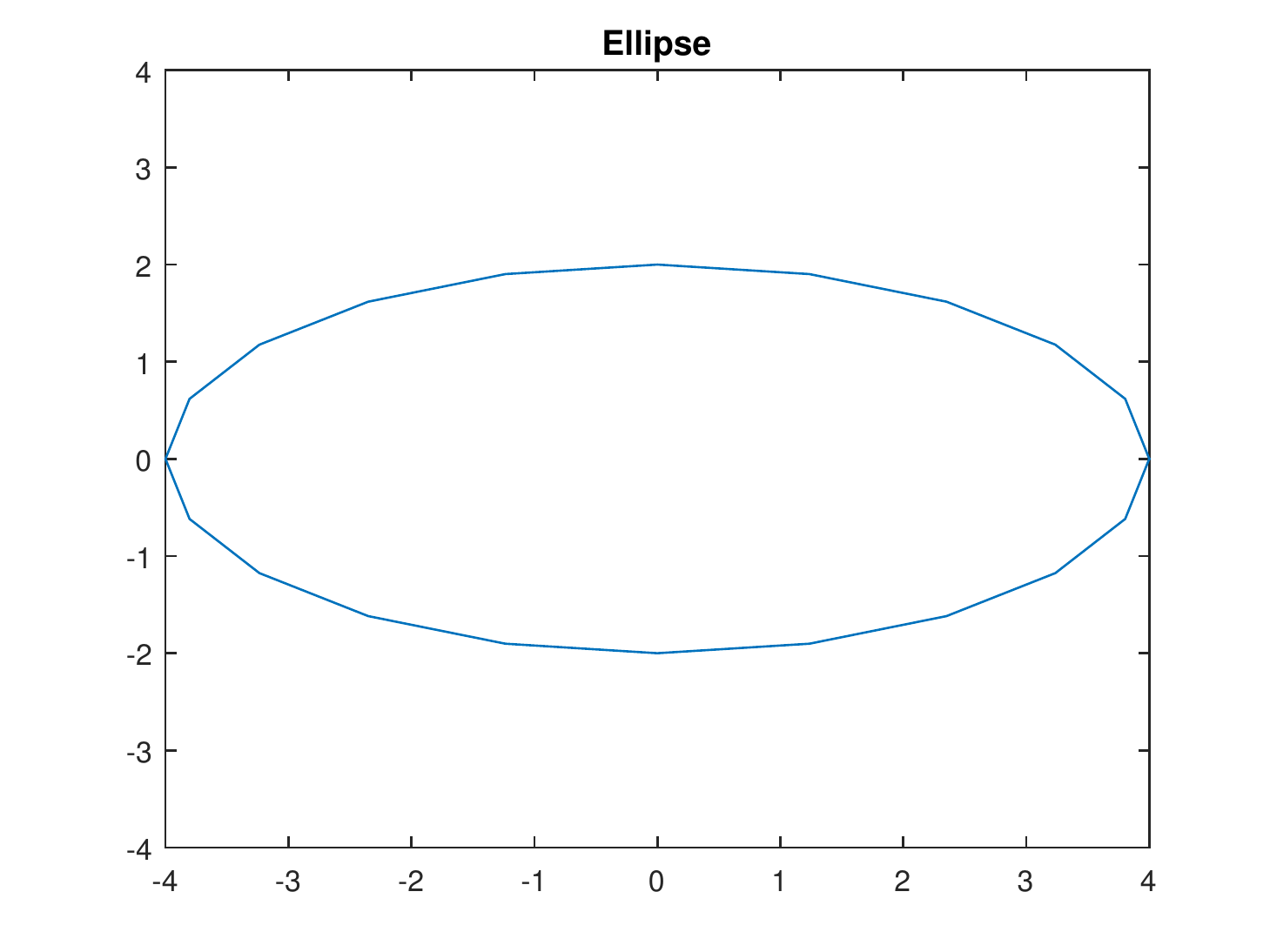}
\includegraphics[scale=.5]{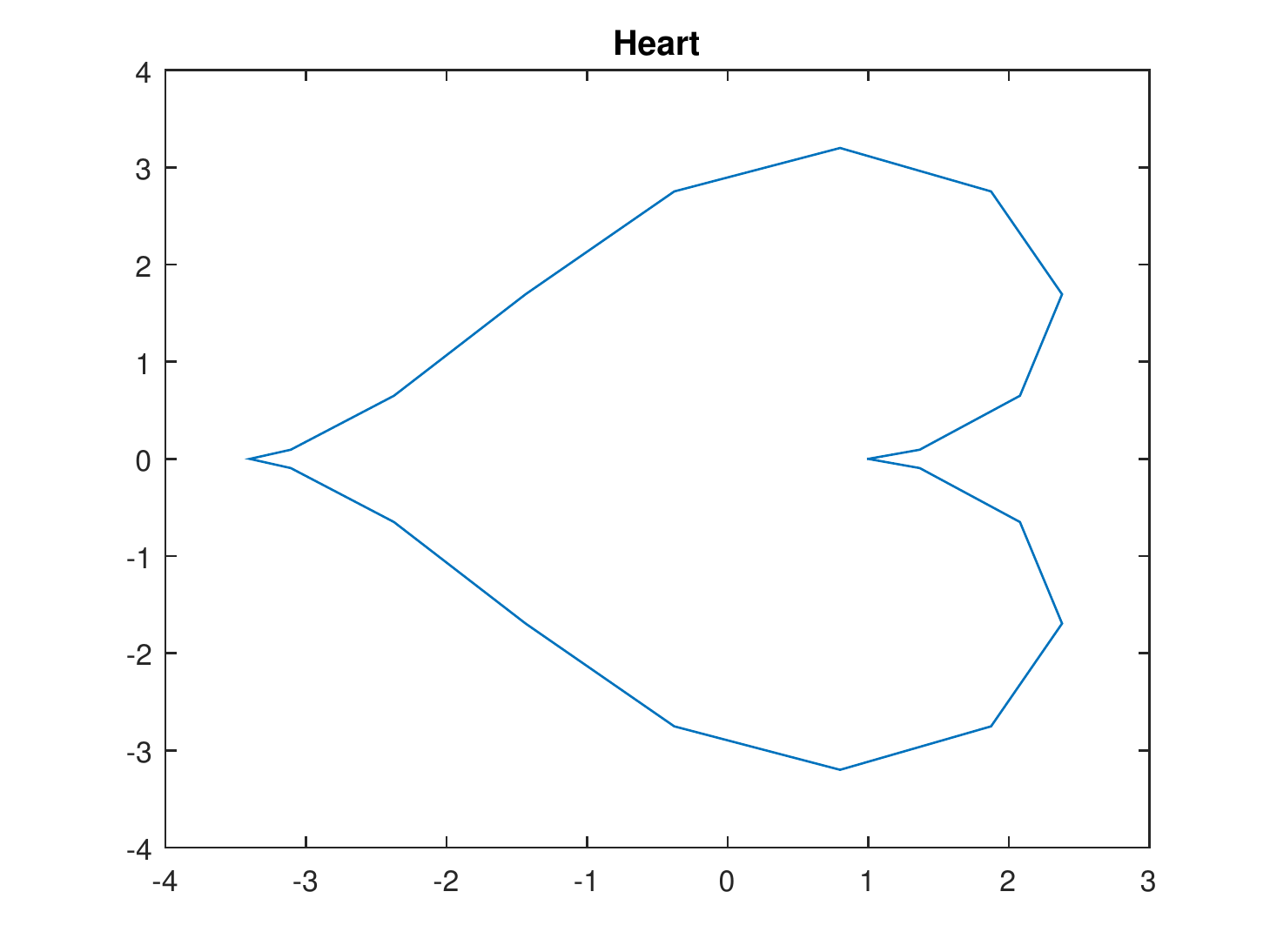}
\caption{A typical ellipse and a typical outlier in the group we consider.}
\label{fig:shapes}
\end{figure}
Figure \ref{fig:shapes} shows a typical ellipse and a typical outlier in this group.  
We show the comparison between the two weighting schemes in Figure \ref{fig:REW}, where $\alpha$ is chosen to be $0.1$, $0.2$, $0.3$, and $0.4$ for different levels of outliers.

It is clear from the figure that the equally weighted scheme absorbs the traits equally from each member of the group; namely when the level of outliers is higher (larger $\alpha$), the shape of the resulting average deviates more from the shapes of the majority, which are a distribution of ellipses. On the contrary, the other weighting scheme gives a robust estimator, in the sense that the obtained average stays close to the majority of the group, in particular when $\alpha$ is large. 

We remark that the use of the weighting schemes depends on the goals of the application. For example, the equally weighted scheme is sensitive to outliers, which can be advantageous if the goal is to to detect outliers in streaming data. On the other hand, if a most representative average is the aim of the application, then the robust algorithm should be used to obtain a robust estimator.

\section{Abnormality Detection}\label{sec:detect}
Once an average for a group is defined and properly calculated, the next two important questions are
\begin{itemize}
\item[(a)] Whether or not new data can be classified into a member of this group by comparing with the average?
\item[(b)] What features from the new data can be used to justify the decision made in (a)?
\end{itemize}

These two questions are part of an even bigger topic of feature extraction in pattern recognition and classification, namely finding features with which data from different categories can be effectively discriminated.  To answer these two questions, we need to seek a set of meaningful variables that can carry necessary information about the nature of the data.  The two variables we choose for the task are (i) a global variable $H$, which is the Hamiltonian that indicates the energy required for the deformation between two sets of landmarks \cite{bib:ckl15}, and (ii) a local variable $\bs P$, which is the collection of momentum at each landmark calculated by Algorithm \ref{alg:average}. 

\begin{figure}[tbh]
\includegraphics[scale=.7]{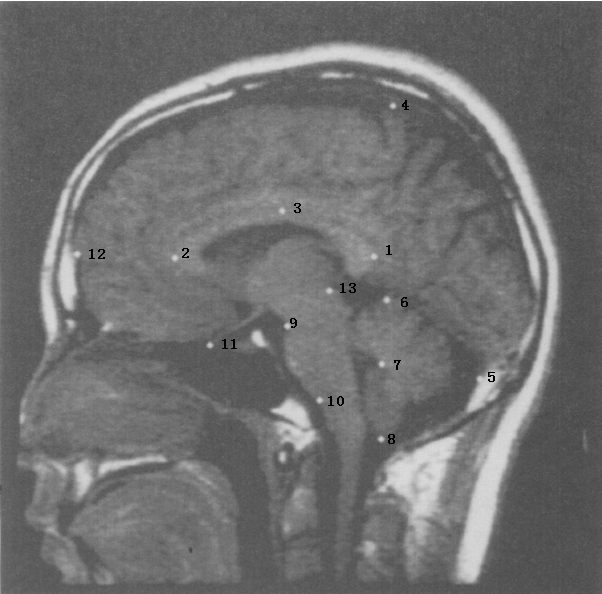}
\caption{Landmark Positions. (1) splenium, posteriormost point on corpus callosum; (2) genu, anteriormost point on corpus callosum; (3) top of corpus callosum, uppermost point on arch of callosum (all three to an
 approximate registration on the diameter of the callosum);
 (4) top of head, a point relaxed from a standard landmark
 along the apparent margin of the dura; (5) tentorium of
 cerebellum at dura; (6) top of cerebellum; (7) tip of
 fourth ventricle; (8) bottom of cerebellum; (9) top of pons, anterior margin; (10) bottom of pons, anterior margin; (11) optic chiasm; (12) frontal pole,
 extension of a line from 1 through 2 until it
 intersects the dura; (13) superior colliculus.}
\label{fig:sample_data}
\end{figure}

The global variable $H$ could be used to distinguish templates of different kinds, for example brains between humans and chimpanzees.
In \cite{bib:ckl15}, we have utilized $H$ for clustering and classifying different simple contour curves. However, it may not be sensitive enough to compare subcategories of the same kind, such as brain structures between people with  and without schizophrenia where only minor local changes could cause a difference. To distinguish such a difference, we need to resort to the local variable.  In this section, we will focus on applying the residual momentum  $\bs P$ from Algorithm \ref{alg:average} for abnormality detection.  A Bayesian predictive approach will be used to assess the information hidden behind the momentum representation $\bs P$. At the same time, we will discuss advantages of using the momentum representation over the traditional location representation. In order to justify this advantage numerically, we use a dataset of  28 parasagittal brain images for illustration. Each image is annotated by 13 landmarks. The dataset can be sorted into two groups. One contains 14 images of non-schizophrenic people, while the other has 14 images from confirmed schizophrenic cases. For more details about this dataset, we refer the reader to the link \cite{bib:link}. The annotated landmarks of a typical brain image in this dataset are shown in Figure \ref{fig:sample_data}.  In applications like this, observations on landmarks of different individuals are essentially close, so we can expect the convergence of Algorithm \ref{alg:average}.

\subsection{Detection}  For each image in the dataset, the annotated landmarks represent locations of human brain structures. It is well known that landmark locations could be different between a schizophrenic and a non-schizophrenic template.  Such differences are sometimes beyond visualization and can only be identified through {mathematical and statistical analysis}. These landmarks carry certain features and information that can help distinguish schizophrenic images from a group of non-schizophrenic templates. We refer to the collection of these landmarks as a predictor. A predictor is viewed as a signature for describing the difference between templates. For the remainder of this section, we focus on finding the landmarks that comprise a predictor for our image data.  The upshot of this exercise could help medical researchers locate those specific landmarks (structures of brain), and shed light on the development of the disease.


It is worth noting that in order to obtain convincing results, more than one method should be applied for validation. One of the traditional Procrustes methods uses superposition of datasets to compare the variations for each landmark. Adopting this approach, Bookstein\cite{bib:Bookstein_schizo} showed that landmarks 1 and 13 are the ``most different'' landmarks between the two groups in the dataset.  For comparison, below we apply Algorithm \ref{alg:average} and use a Bayesian approach on momentum coordinates of each template to detect the abnormal landmarks. 

Our algorithm begins with finding the momentum coordinates for each template in the database. Momentum coordinates $\{P^N_i\}_{i=1}^{14}$ for templates $\{I^N_i\}_{i=1}^{14}$ of the non-schizophrenic group will be calculated through Algorithm \ref{alg:average}. For each $i$, $P^N_i$ is a 13 by 2 matrix, in which the $j$th row ($j=1,2,\cdots,13$) contains the momentum coordinate calculated at the $j$th landmark. We denote the resulting average of this group by $\overline{I}_N$.  After $\overline{I}_N$ is found, momentum representations based on the average calculated from the previous step for each individual in the schizophrenic group, denoted $\{P^S_i\}_{i=1}^{14}$, will be computed by the Riemannian logarithm map, i.e. $P^S_i = \text{Log}_{\overline{I}_N}(I^S_i)$.  To apply statistical methods for the momentum coordinates, we construct the observation data matrix for each landmark. For each landmark $j=1,2,\cdots,13$, we collect the $j^{\text{th}}$ row of the momentum coordinates $\{P^N_i\}_{i=1}^{14}$ of the non-schizophrenic group to form a $14\times 2$ matrix  $N_j$. Similarly, we construct the data matrix $S_j$ for the schizophrenic group.
\begin{note}
We remark that for the $k^{th}$ image in each group, e.g. the non-schizophrenic group, the momentum coordinators $N_1^{k},\cdots\, N_{13}^{k}$ are correlated, whereas  $N_1^{k}$ is independent of $N_1^{l}$, the momentum coordinate of first landmark of the $l^{th}$ image in this group. Our statistics in the next section are based on the independent momentum coordinates computed, respectively, from the 14 different images with respective to the average within the two groups. 
\end{note}


\subsection{Statistical methods for momentum data}


%
%
%


In this section, we use two statistical methods to process the momentum data  $\{N_j, S_j \}$ for landmarks $j=1, 2,\cdots, 13$. Our aim is to identify those landmarks associated with changes between non-schizophrenia group and the compared schizophrenia group through analysis on their residual momentum representation. If both methods give consistent conclusions that are also consistent with the classical Procrustes method, the result by our method will be trustworthy in this task.  We do not use classical hypothesis testing since our sample sizes are relatively small and most of the classical results rely on large sample asymptotics. Thus, we use a simple distance-based approach and also construct a Bayesian predictive distribution which conditions on the data and does not rely on asymptotics.

\subsubsection{Method 1(Norm of mean momentum)}


The first statistical method we use to analyze the momentum data is to directly compare the norm of mean momentum at each landmark. This method follows an intuitive idea that landmarks with large mean momentum calculated from the schizophrenic group may indicate abnormalities. In this method, for each $N_j$ and $S_j$, $j=1,\cdots,13$, we compute the $l_2$ norm of mean momentum by 
\beq
||\overline{E}(N_j)||=\parallel\frac{1}{14}\sum_{k=1}^{14}N_j^{k}\parallel_2.
\eeq
Note that for each $N_j^k$, there are two components. The $l_2$ norm evaluates the square root of the sum of component squares.  Table \ref{tab:meanPN} and Table \ref{tab:meanPS} summarize the $l_2$ norm of the mean momenta calculated at each landmark for the two groups, respectively. For the schizophrenic group (see Table \ref{tab:meanPS} ), we note that Landmarks 1, 6 and 13 have unusually large values compared to other landmarks. We group these landmarks together as a predictor for this method.
\begin{table}[th]
	\begin{tabular}{|c|c|c|c|c|c|c|c|c|c|c|c|c|c|}
		\hline
		\hline
		\multicolumn{8}{ |c| }{Norm of Mean Momentum for N Group}\\
		\hline
		Landmark & 1 & 2 & 3 & 4 & 5 & 6 & 7\\  
		\hline
		$||\overline{E}(N_j)||$        & 0.0001 & 9.39e-06 & 	1.17e-05 & 9.07e-06 & 1.11e-05 & 0.0001 & 3.74e-05 \\
		\hline
		\hline
		Landmark & 8 & 9 & 10 & 11 & 12& 13\\
		\hline
		$||\overline{E}(N_j)||$ & 2.6610e-05 & 1.3453e-05 & 1.9547e-05 & 2.4342e-05 & 4.8575e-06&5.9788e-05\\
		\hline
		\hline
	\end{tabular}
	\caption{$l_2$ norm of the mean momentum at each landmark in non-scizophrenic group. } \label{tab:meanPN}
\end{table}	
\begin{table}[th]	
	\begin{tabular}{|c|c|c|c|c|c|c|c|c|c|c|c|c|c|}
		\hline
		\hline
		\multicolumn{8}{ |c| }{Norm of Mean Momentum for S Group}\\
		\hline
		Landmark & 1 & 2 & 3 & 4 & 5 & 6 & 7\\  
		\hline
	$||\overline{E}(S_j)||$ & $\bs {0.2231}$ & 0.0338 & 0.0533 & 0.0995 & 0.0932 & $\bs{0.1391}$ & 0.0432 \\
		\hline
		\hline
		Landmark & 8 & 9 & 10 & 11 & 12& 13\\
		\hline
		$||\overline{E}(S_j)||$ & 0.0722 & 0.0635 & 0.0270 & 0.0826 & 0.0873& $\bs{0.2182}$\\
		\hline
		\hline
	\end{tabular}
	\caption{$l_2$ norm of the mean momentum at each landmark in schizophrenic group. Landmark 1, 6 and 13 have unusually large values.}
		\label{tab:meanPS}
\end{table}


\subsubsection{Method 2 (Bayesian Posterior Predictive distributions)}

While the method by observing the sample mean can help provide certain information on abnormal location of landmarks,
using Bayesian (posterior) predictive distributions we can obtain distributions for the landmarks in each of the two groups. This offers the advantage that the inference is not on the means of the landmarks, but on the distributions of the landmarks themselves. The landmarks are `observables' while their means are parameters and `latent'; inference on observables can be validated with future data as more images are obtained. Predictive distributions of the momentum data for each landmark are obtained via a Gibbs sampler for sampling from the posterior predictive distributions; details of the statistical modeling and its implementation are in Appendix \ref{sec:method3}.

\begin{figure}[tbh]
 	\subfigure[Landmark 1]{\includegraphics[scale=.2]{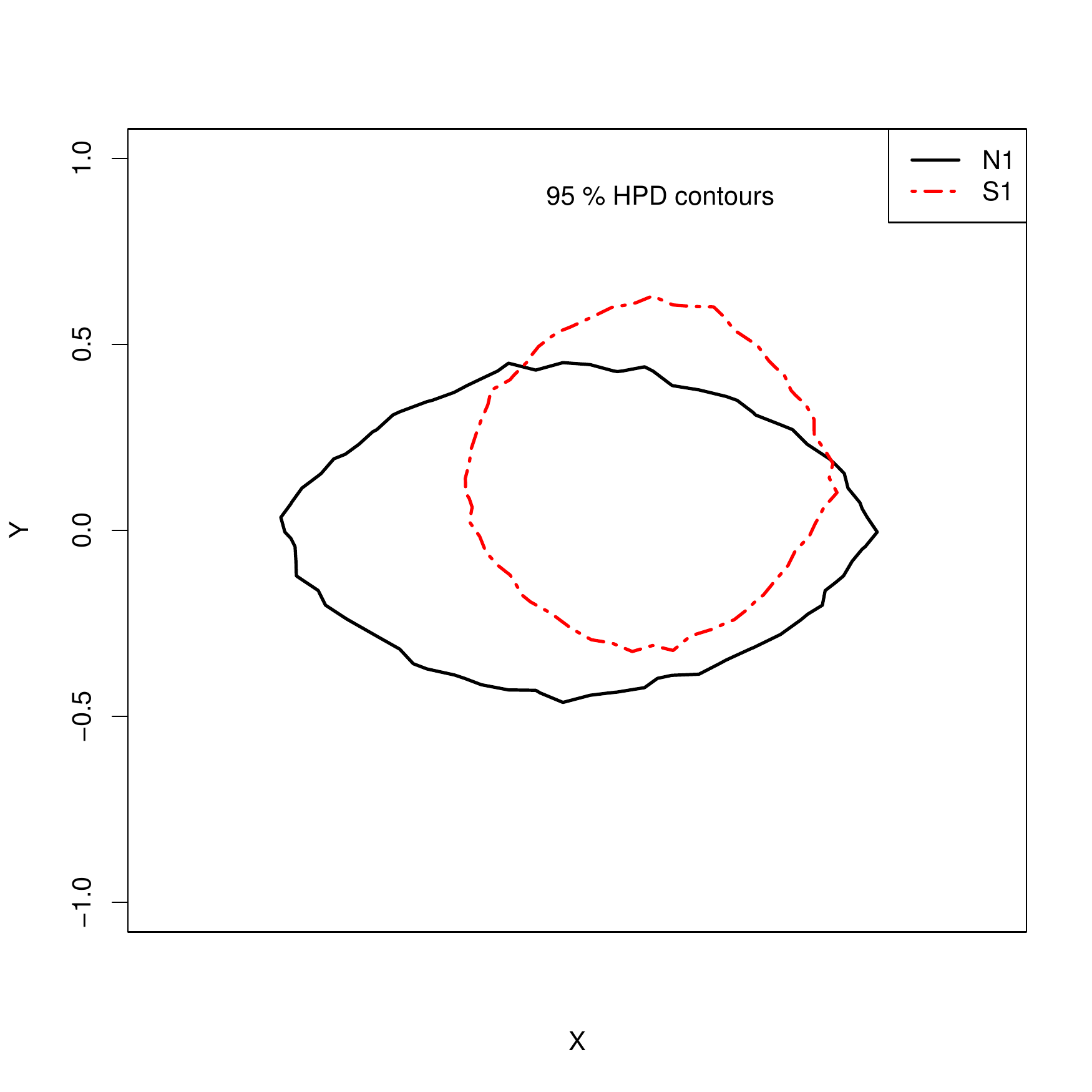}}
 	\subfigure[Landmark 2]{\includegraphics[scale=.2]{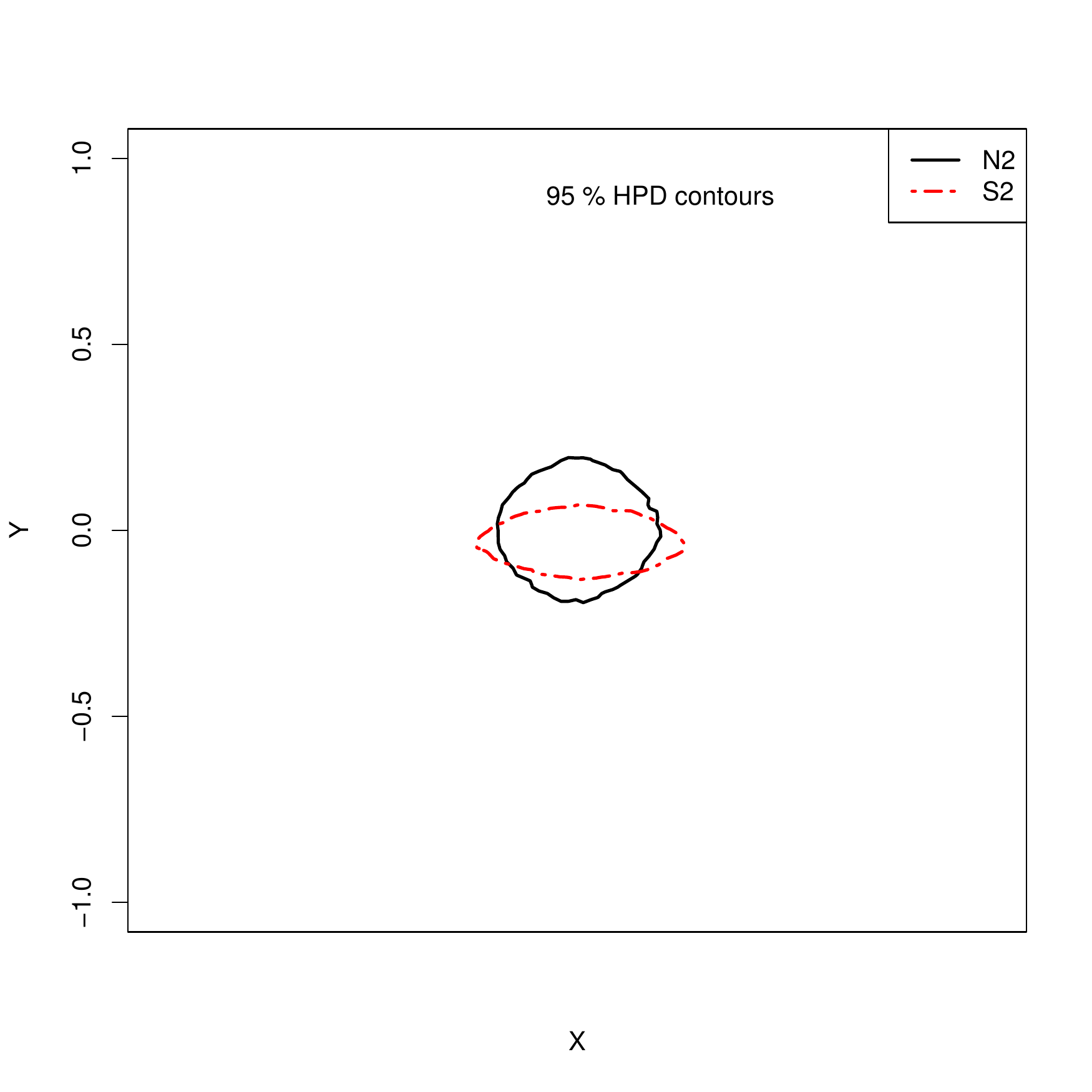}}
 	\subfigure[Landmark 3]{\includegraphics[scale=.2]{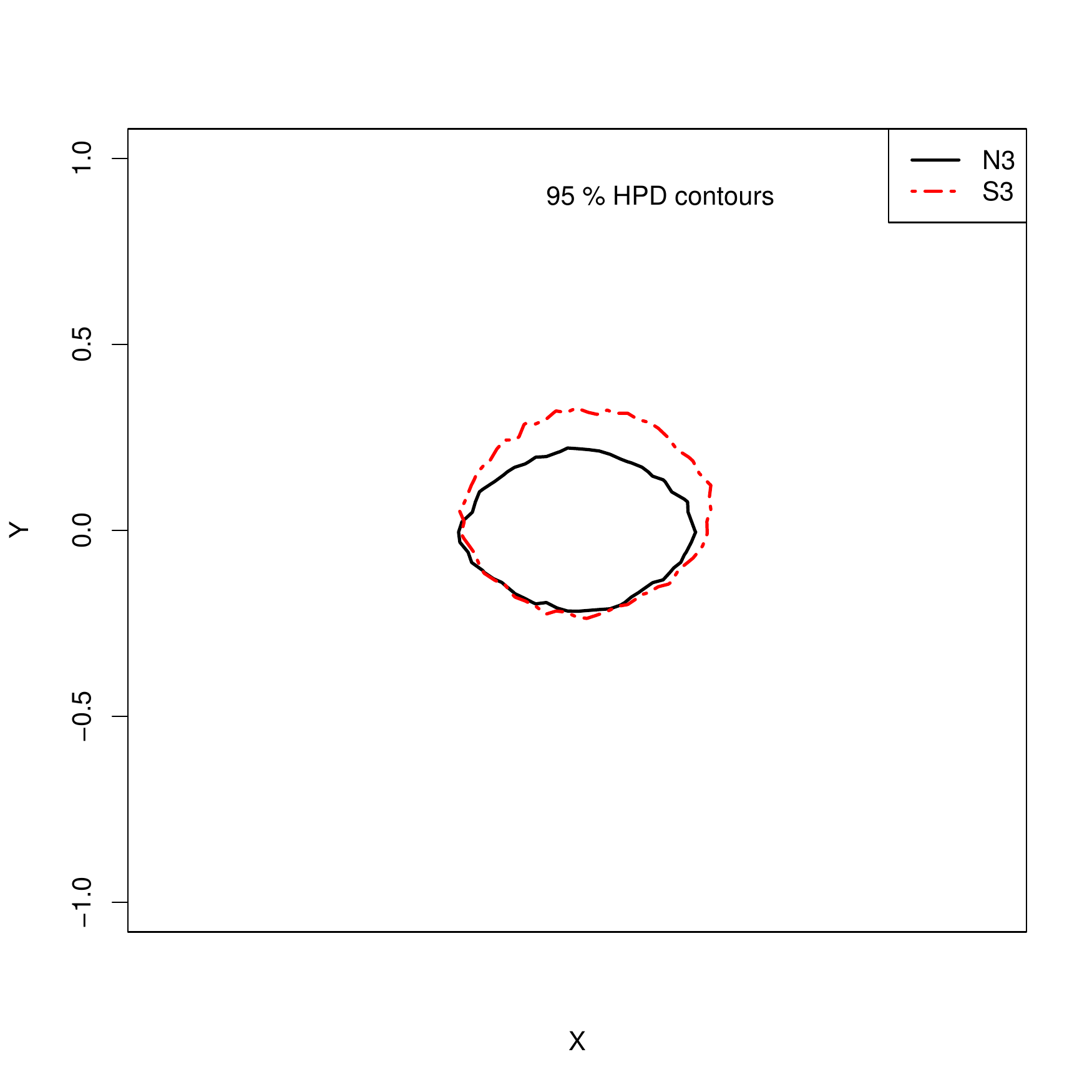}}
 	\subfigure[Landmark 4]{\includegraphics[scale=.2]{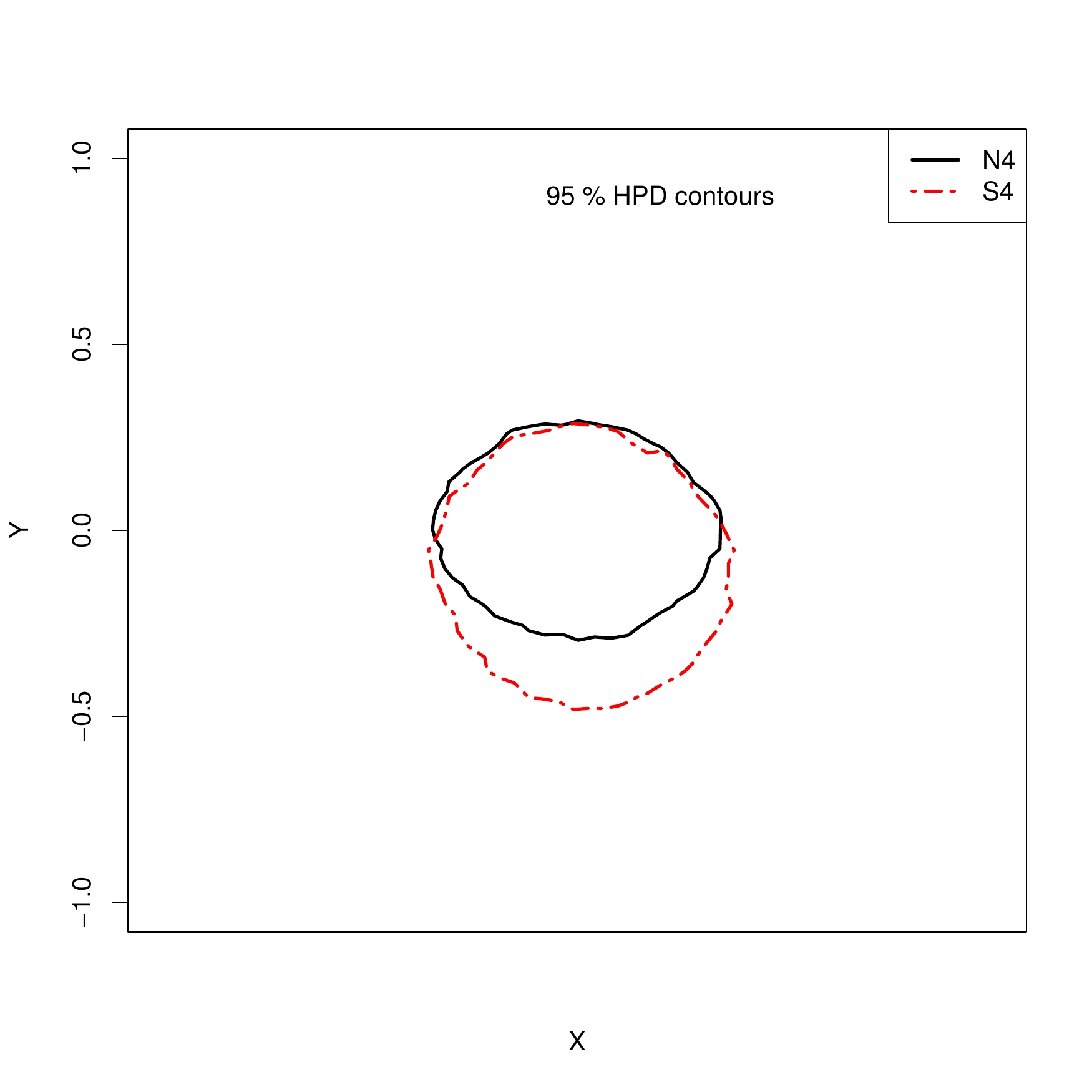}}
 	\subfigure[Landmark 5]{\includegraphics[scale=.2]{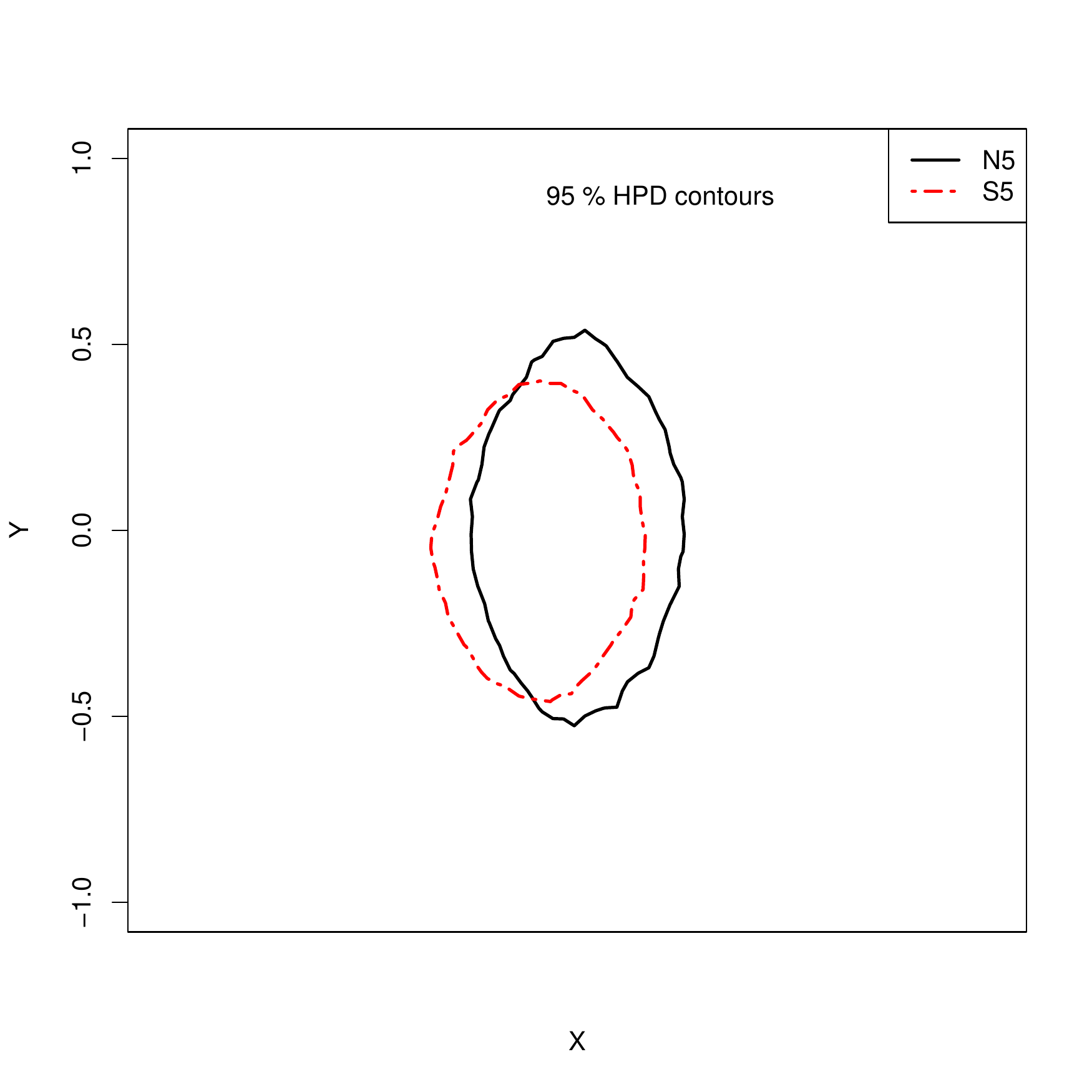}}
 	\subfigure[Landmark 6]{\includegraphics[scale=.2]{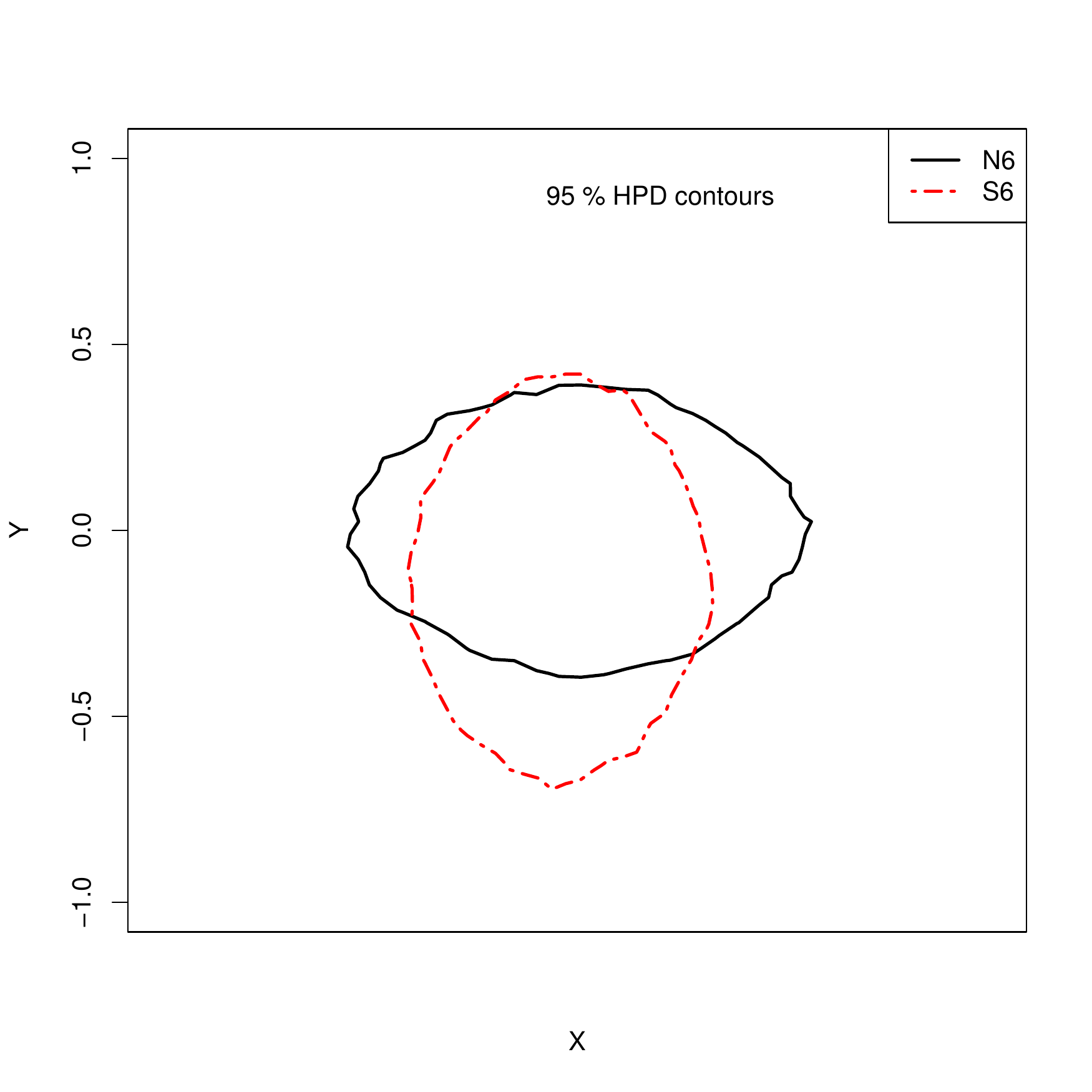}}
 	\subfigure[Landmark 7]{\includegraphics[scale=.2]{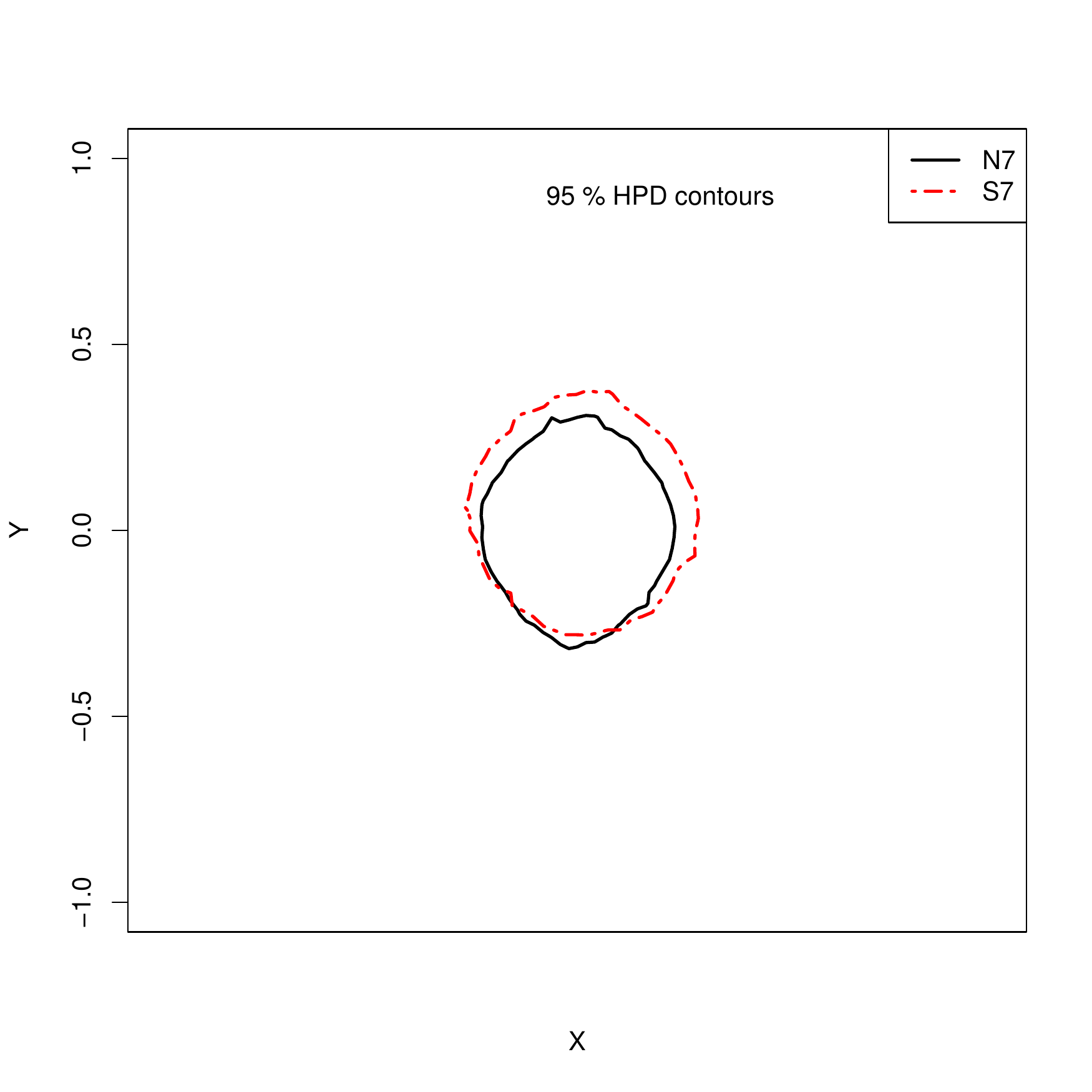}}
 	\subfigure[Landmark 8]{\includegraphics[scale=.2]{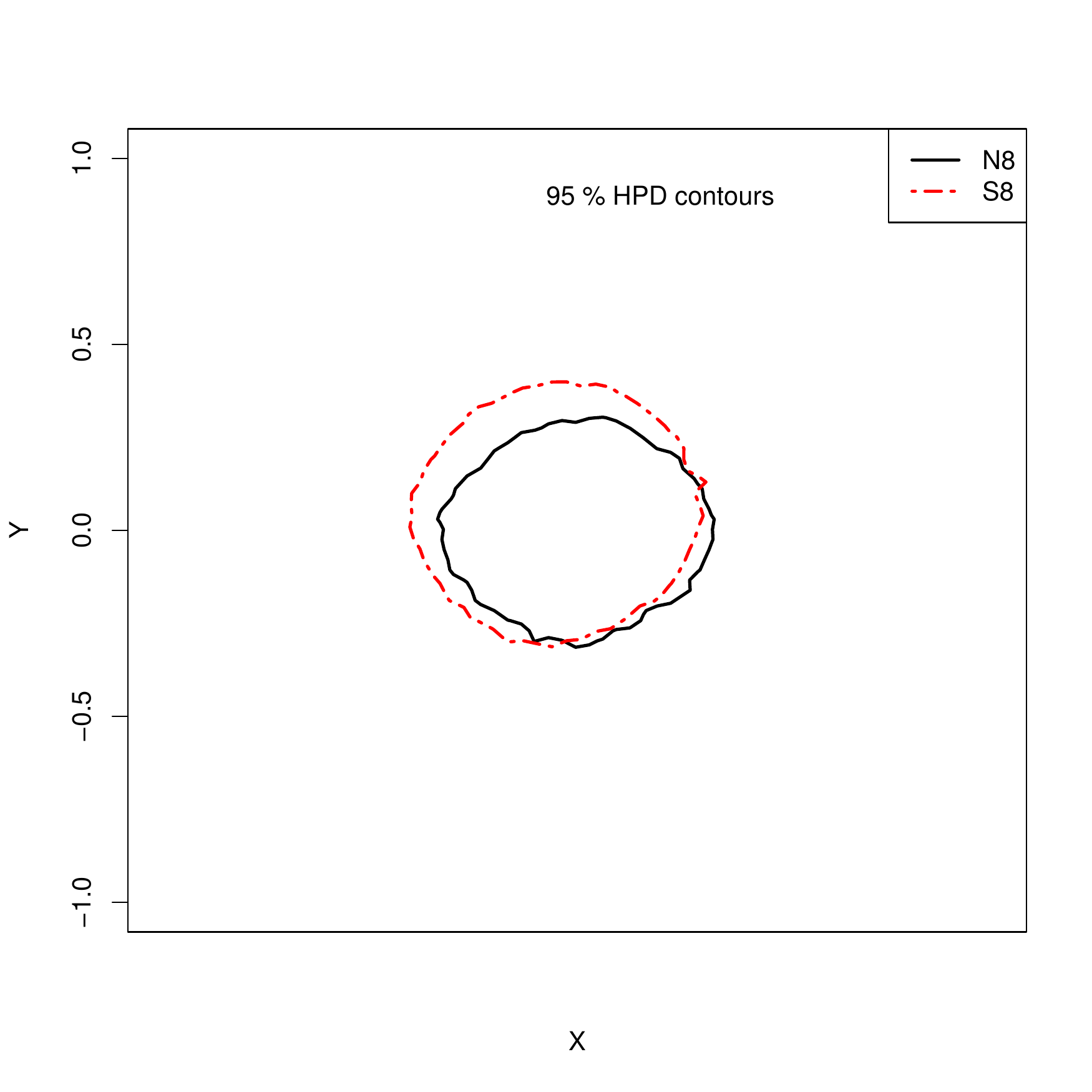}}
 	\subfigure[Landmark 9]{\includegraphics[scale=.2]{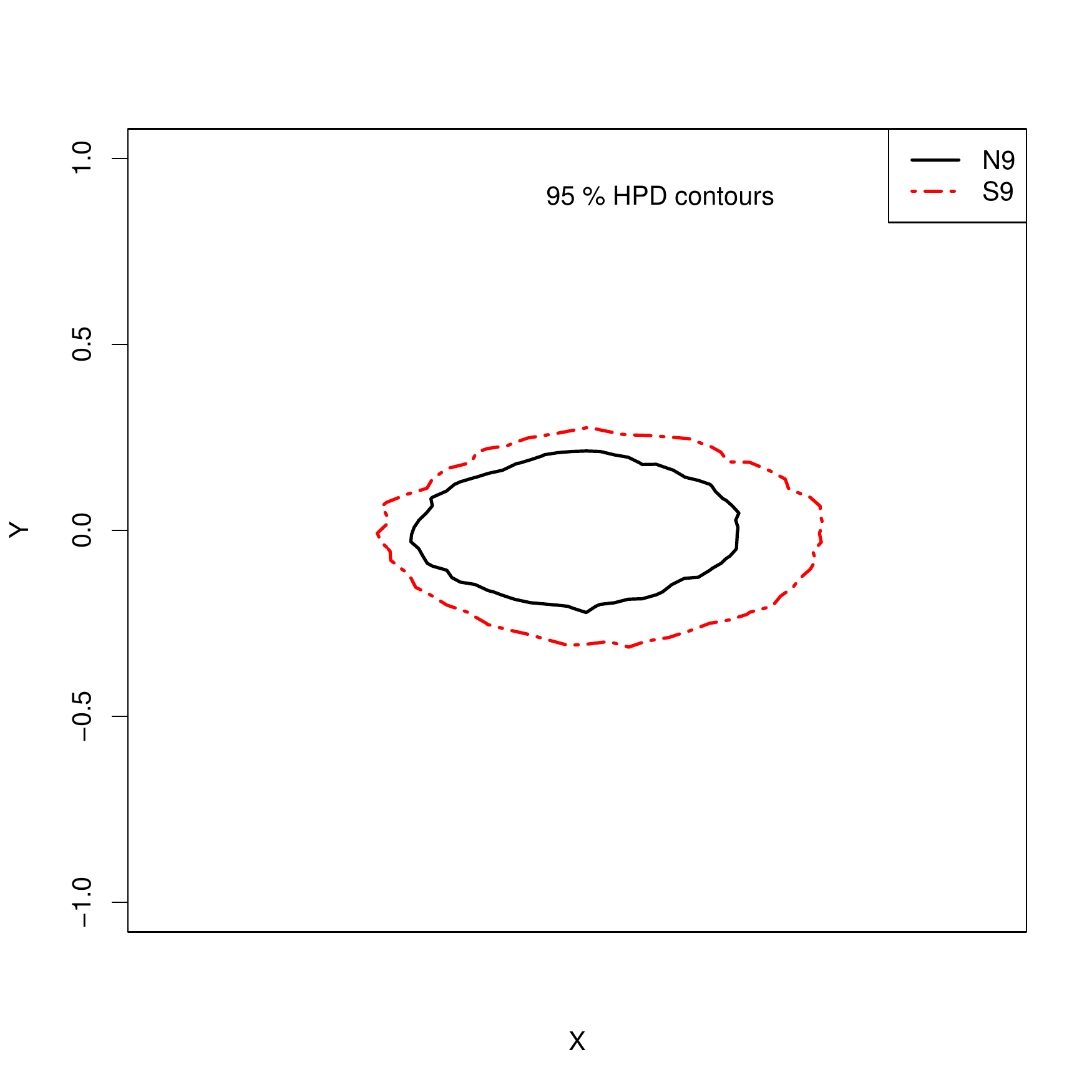}}
 	\subfigure[Landmark 10]{\includegraphics[scale=.2]{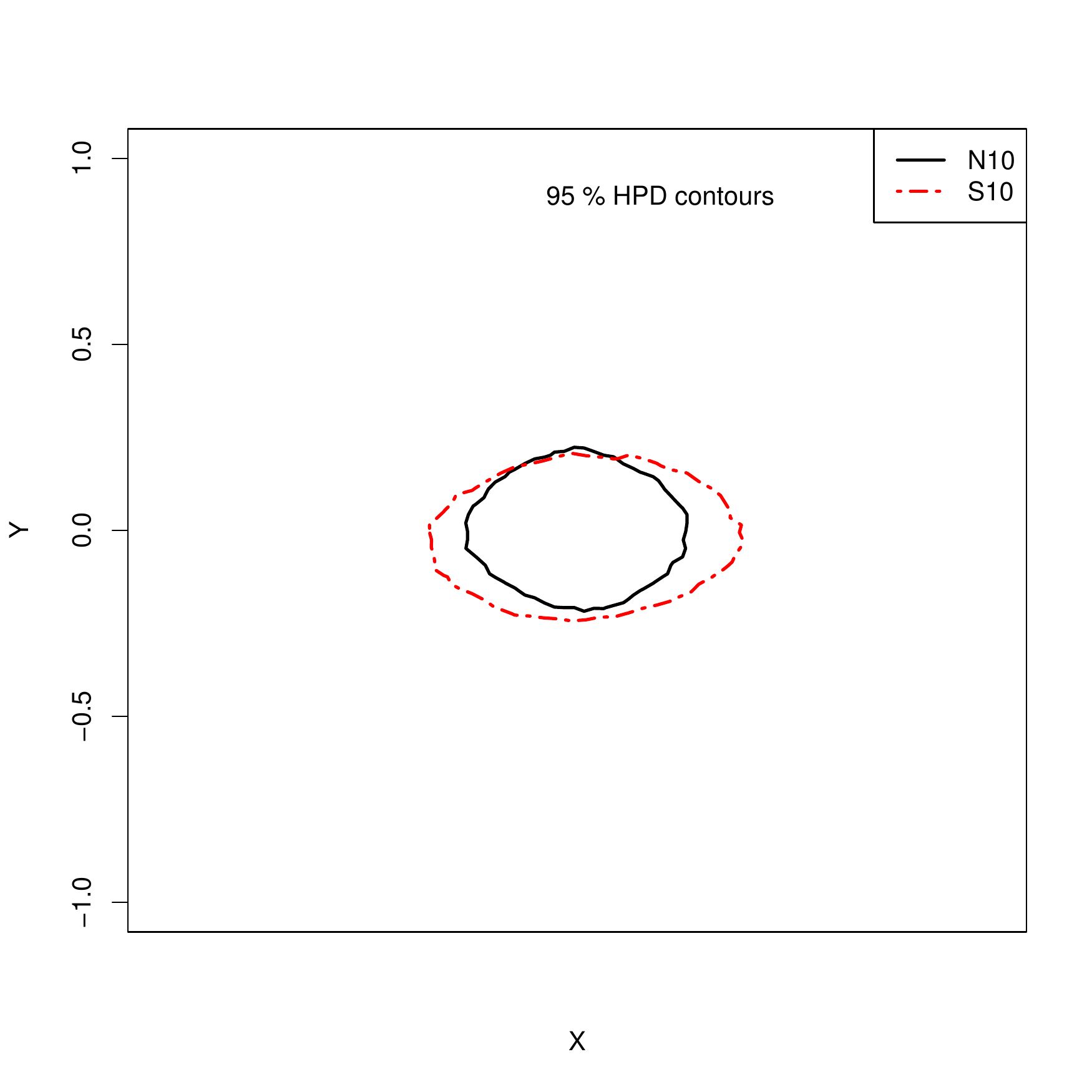}}
 	\subfigure[Landmark 11]{\includegraphics[scale=.2]{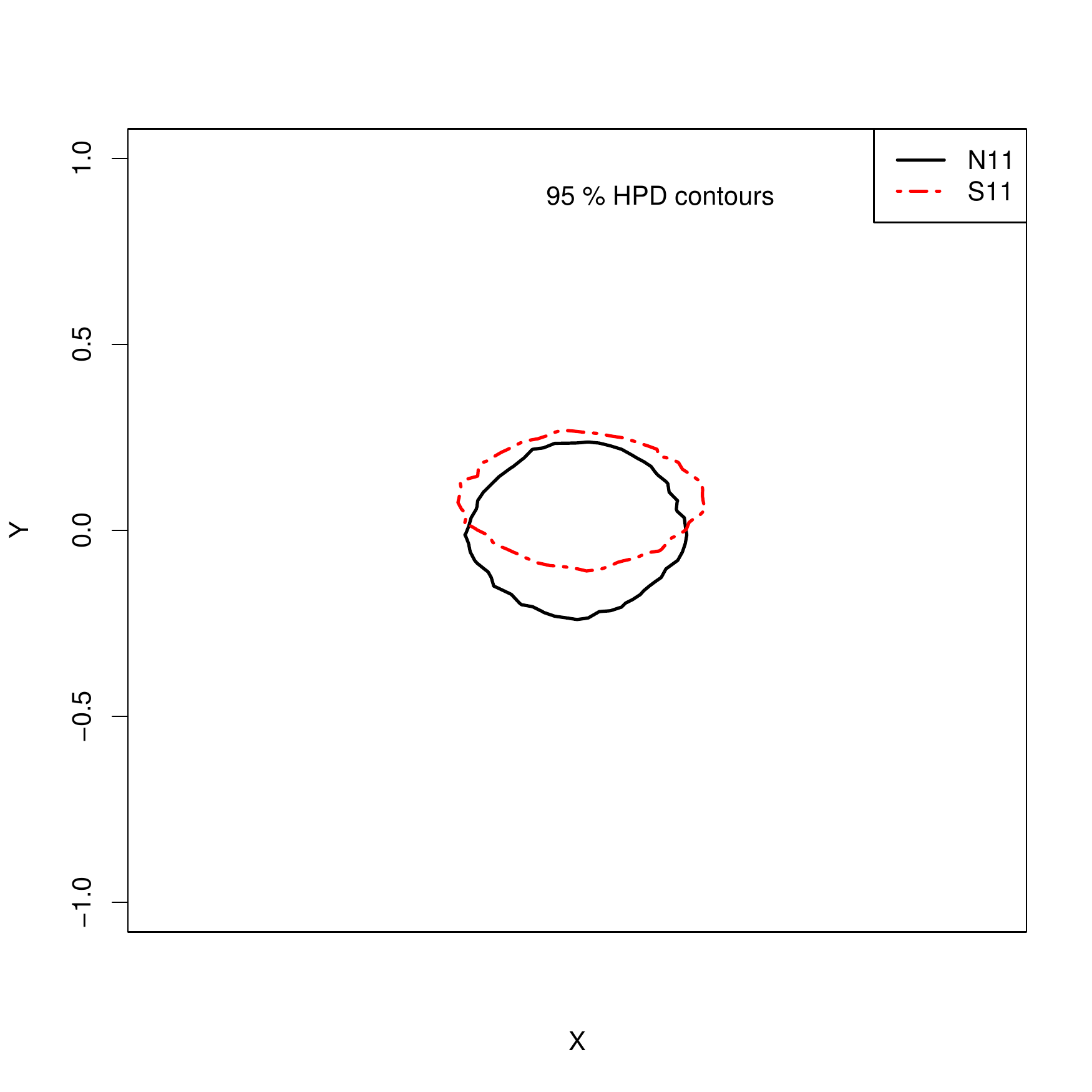}}
 	\subfigure[Landmark 12]{\includegraphics[scale=.2]{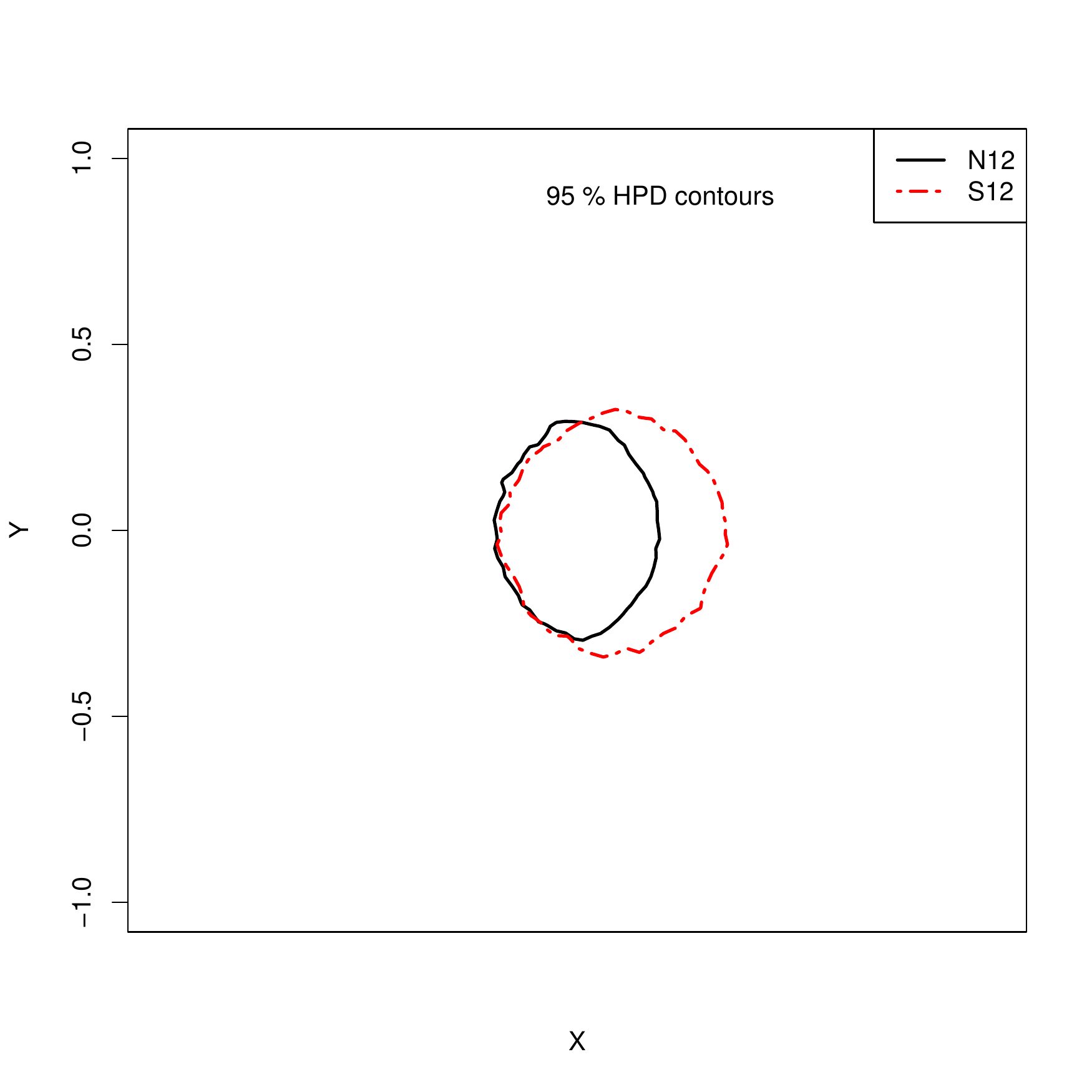}}
 	\subfigure[Landmark 13]{\includegraphics[scale=.2]{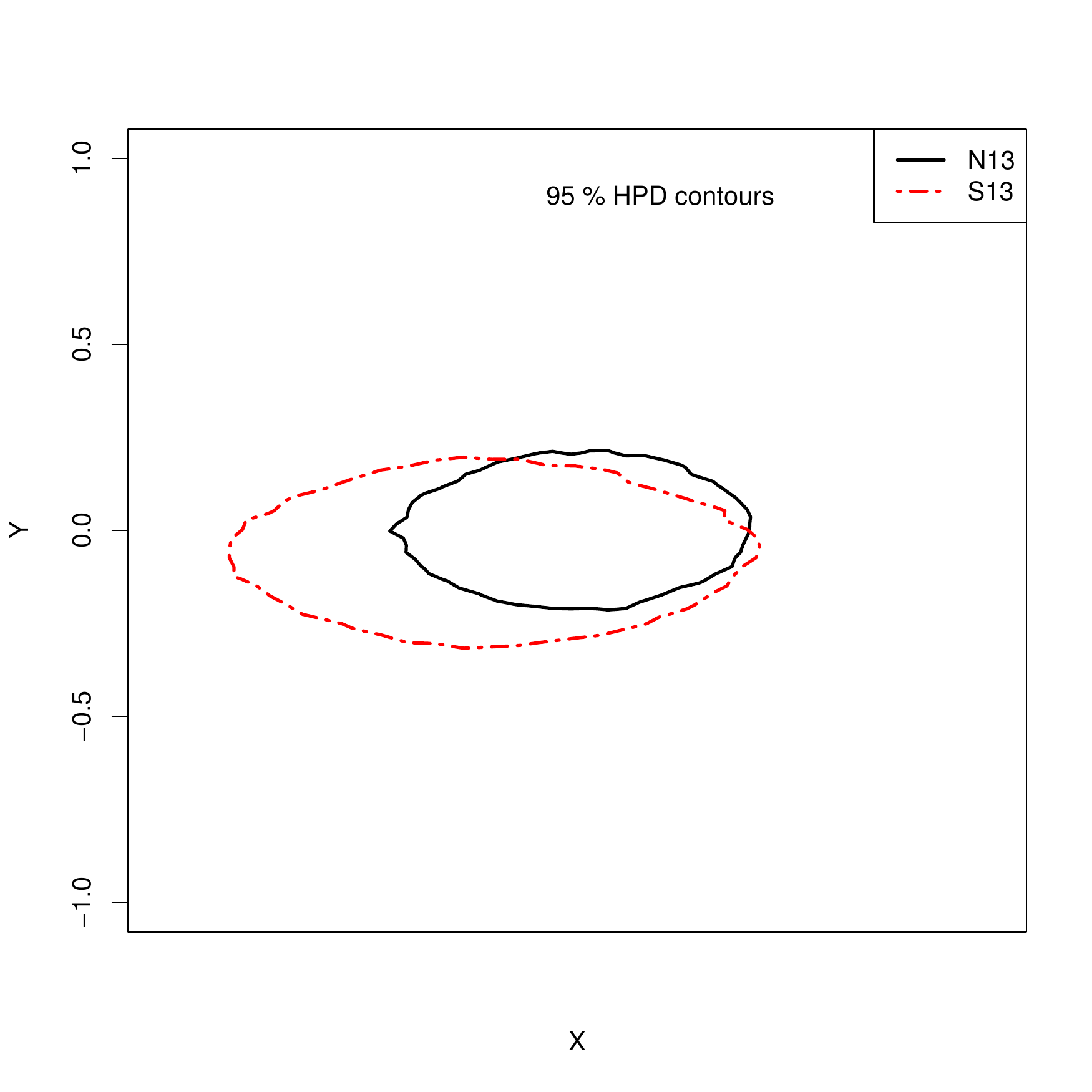}}
 	\caption{95\% contour plots of the predictive distribution for each of the 13 landmarks. Black solid line is calculated from N group, Red dashed line is calculated from S group. Landmarks with their N and S contour plots differing much will be picked out as a predictor.}
 	\label{fig:contour}
 \end{figure}

For each landmark, the respective predictive distributions of the two groups are compared to decide whether there is a significant difference. The predictive distribution $z=p(x,y|\{x_i,y_i\}_{i=1}^{14})$ is a surface, which is hard for us to analyze the difference. Instead, we use the contour lines of $p(x,y|\{x_i,y_i\}_{i=1}^{14})=0.95$ for their comparison. Figure \ref{fig:contour} shows the plots of the 95\% contour lines of the two groups for each of the 13 landmarks.

To quantify the difference between the contour lines for each landmark, we compute the ratio of the overlapping area of the two contour regions over the union of the two regions. Let $\Omega ^N_i$ and $\Omega ^S_i$ represent the contour region for each group, respectively,  for the $i^{\text{th}}$ landmark. We define the ratio as $r_i := \frac{|\Omega^N_i \cap \,\Omega^S_i|}{|\Omega^N_i \cup \,\Omega^S_i|}$. We expect that the smaller the value $r_i $ is, the more different the two distributions will be. In practice, we can approximate $r_i $ by  $\hat{r}_i $ to avoid the extra computation of 2-D integration, where $\hat{r}_i := \frac{B^N_i \cap \, B^S_i}{B^N_i \cup \, B^S_i}$. Here $B^N_i := X^N_i  \times Y^N_i$ with $X^N_i $ and $ Y^N_i$  are the central $95\%$ intervals of the marginal predictive distributions. $B^S_i$ is defined exactly the same way. 
 

Table \ref{table:olap} shows the approximated ratios for the 13 landmarks. Landmarks with $r<0.5$ (i.e. approximately less than a half of the total region is overlapped) is considered as having significant difference between the two groups, and they will be grouped together as a predictor.  In this case, the predictor for this method includes landmarks 1, 2, 6 and 13. We remark that the momentum field used in our calculations is advantageous. For comparison,  we repeat exactly the same calculations, but replace the momentum data by the original position data.  Table \ref{table:olap_p} is the result of using position data, the counter part to Table \ref{table:olap}. As we can see that no landmarks have ratios that are less than 0.5 in Table \ref{table:olap_p}. This indicates that the momentum representation has better separability than the location one. 


\begin{table}
\begin{tabular}{|c|c|c|c|c|c|c|c|c|c|c|c|c|c|}
\hline
\hline
\multicolumn{8}{ |c| }{Results on Momentum Data}\\
\hline
Landmark & 1 & 2 & 3 & 4 & 5 & 6 & 7\\  
\hline
$r$        & $ \bf 0.4535$ & $\bf 0.4286$ & 0.6906 & 0.6542 & 0.5561 & $\bf 0.4972$ & 0.7187 \\
\hline
\hline
Landmark & 8 & 9 & 10 & 11 & 12& 13\\
\hline
$r$ & 0.6825 & 0.5348 & 0.6155 & 0.5470 & 0.5554& $\bf 0.4538$\\
\hline
\hline
\end{tabular}
\caption{Overlapped region rate for each landmark using the generated marginal distribution from the extracted momentum data. Landmark 1, 2, 6 and 13 got separated from others for their overlapped rate $r<0.5$. They could be used as a predictor with their momentum representation.} \label{table:olap}
\end{table}

\begin{table}
\begin{tabular}{|c|c|c|c|c|c|c|c|c|c|c|c|c|c|}
\hline
\hline
\multicolumn{8}{ |c| }{Results on Position Data}\\
\hline
Landmark & 1 & 2 & 3 & 4 & 5 & 6 & 7\\  
\hline
$r$        & 0.7900 & 0.6636 & 0.5480 & 0.6249 & 0.5559 & 0.6045 & 0.5528 \\
\hline
\hline
Landmark & 8 & 9 & 10 & 11 & 12& 13\\
\hline
$r$ & 0.5887 & 0.5022 & 0.6171 & 0.5560 & 0.6066& 0.6063\\
\hline
\hline
\end{tabular}
\caption{Overlapped region rate for each landmark using the generated marginal distribution from the original location data. No landmarks have $r<0.5$, meaning the distribution are quite similar between the two group of data. The potential difference is not well reflected in the location representation.}\label{table:olap_p}
\end{table}
The above three methods provide different viewpoints for our investigation. Method 2 finds a predictor by directly examining  the peak values of the parameters (the 2-norm of mean momentum in our case). 
Method 3 uses Bayesian inference to search for a predictor. 
By examining the mean momentum, we found an additional abnormal landmark 6 when compared to Procrustes Method. The Bayesian approach, which also models the data, helps to dig out another suspected abnormal landmark 2. The three methods here seem to discover abnormal landmarks at different scales. In Table \ref{table:predictor}, we summarize the predictors found in the literature and our methods. The results are consistent. 


\begin{table}
\begin{tabular}[t]{|c|c|}
\hline
\hline
Method & Predictor\\
\hline
Procrustes Method in \cite{bib:Bookstein_schizo} & 1,13\\
\hline
Mean Momentum (MM)& 1, \textbf{6},13\\
\hline
Predictive Distribution Contour (PDC) & 1, \textbf{2}, \textbf{6}, 13\\
\hline
\end{tabular}
\caption{Summary of predictors from different methods.}
\label{table:predictor}
\end{table}

\begin{figure}[tbh]
	\includegraphics[scale=0.7]{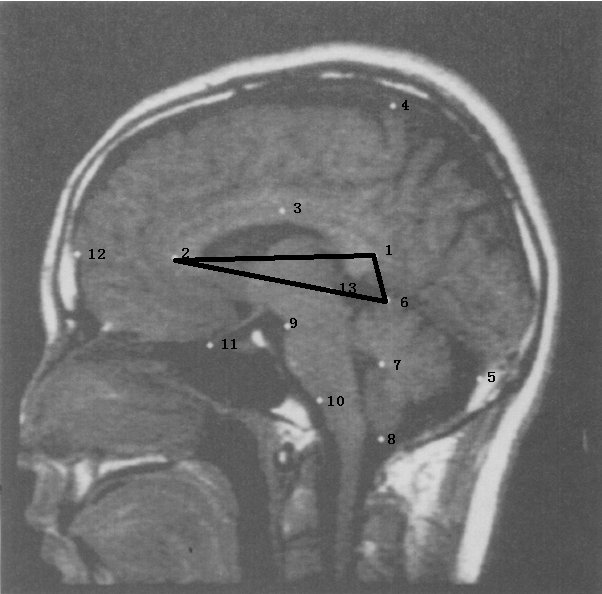}
	\caption{Region enclosed by detectors 1,2,6,13. Our calculation based on sample data suggests physical changes at mid-brain accompany schizophrenia. Certain important structures with in this region such as: Thalamus, Hippocampus and Third Ventricle may be worth investigated. Some medical results \cite{bib:med01, bib:med02} seems to support this observation.}\label{fig:enclose}
\end{figure}

Finally, we comment that from Table \ref{table:predictor}, the structure differences between the groups of schizophrenia and non-schizophrenia in our database most likely occur at the locations annotated by landmarks 1, 2, 6 and 13. Indeed, the region enclosed by these landmarks (see Figure \ref{fig:enclose}) were investigated in clinical trials and found to be relevant to schizophrenia from some of the literature \cite{bib:med01, bib:med02}. 

\section{Conclusions}

We have proposed a Bayesian approach using the residual momentum representation of landmark templates through the calculation of a group average. We prove a convergence theorem for the averaging algorithm with different weights under certain conditions. The resulting residual momentum variable from the averaging process provides a local template-based linear coordinate system. It also serves as a representation which could enhance potential differences between groups under comparison when using the original location representation cannot, and thus is advantageous for related statistical analysis like abnormality detection. We apply the proposed approach on a medical image database. The database contains 14 non-schizophrenic and 14 schizophrenic templates annotated with 13 landmarks. The region of brain structure changes in schizophrenia found in our study is highly coherent with that in the literature. 

Finally, though there are certainly much remain to do, the applications of this proposed algorithms could have impacts on clinical medicine, forensic, and surveillance. For example, medical image abnormality detection will help physicians for early diagnoses on schizophrenia, alzheimer, and heart diseases. Fast automatic face recognition algorithms will provide forensic identification and instantaneous image matching in surveillance that helps governments for intelligence gathering, the prevention of crime, the protection of a process, person, group or object, or for the investigation of crime.



%



\appendix

\section{Statistical model specifics and implementation of method 2}\label{sec:method3}

The data are computed from the analysis of images from two groups of 14 non-schizophrenic and 14 schizophrenic brains, so that within each group at each of 13 landmarks, we have 14 observations (number of brains in group) on bivariate data (x,y components). Notationally, the vectors in the $x$ and $y$ directions result in data $\{(x,y)^{\prime}_{i,j,k}; i=1, \ldots, 14, j=1, \ldots, 13, k=1,2 \}$ which we denote as $({\bf X, Y})^{\prime}$. In what follows, we use the notation $\{(x,y)^{\prime}\}_{(l)}$ to denote a subset of $({\bf X, Y})^{\prime}$ formed by collecting vectors over the index $l$.
 
For fixed $(j,k)$, we model the vectors $\{ (x,y)^{\prime}_{i,j,k}\}_{(i)}$ as conditionally independent bivariate normal random variables given the mean, ${\boldsymbol \mu}_{j,k} = (\mu_{x,j,k}, \mu_{y,j,k})^{\prime}$, and covariance matrix ${\boldsymbol \Sigma}_{j,k}$, yielding the joint probability distribution,
\begin{equation} \label{bvn}
[\{(x,y)^{\prime}_{i,j,k}\}_{(i)}|\boldsymbol{\mu}_{j,k}, \boldsymbol{\Sigma}_{j,k}] =  \prod_{i=1}^{14} N((x,y)^{\prime}_{i,j,k} | {\boldsymbol \mu}_{j,k},{\boldsymbol \Sigma}_{j,k})
\end{equation}
where $[.]$ denotes a probability distribution with $[.|.]$ for the conditional version.  For purposes of data modeling, the bivariate distribution can be rewritten as the product of  $[x_{i,j,k}]$, the marginal distribution of the first component of the $(x,y)^{\prime}_{i,j,k}$ vector, with  $[y_{i,j,k}|x_{i,j,k}]$, the conditional distribution of the second component given the first, where both are univariate normal distributions. This reformulation yields a likelihood where the covariance structure is a function of the two standard deviations ($\sigma_{x,j,k}, \sigma_{y,j,k}$) and one correlation ($\rho_{j,k}$). There is more flexibility in assigning priors to scalar parameters (Lunn et al. 2013, p.226), whereas the usual bivariate formulation restricts priors for the covariance matrix, ${\boldsymbol \Sigma}$, to the Wishart distribution with little flexibility in the prior's shape.
Given data from a $(j,k)$ combination, we denote the contribution to the likelihood for ${\boldsymbol \theta}_{j,k} = (\mu_{x,j,k}, \mu_{y,j,k}, \sigma_{x,j,k}, \sigma_{y,j,k}, \rho_{j,k})$ as  $L({\boldsymbol \theta}_{jk} | \{ (x,y)^{\prime}_{i,j,k}  \}_{(i)} )$, which is proportional to the quantity in equation (\ref{bvn}). Then, the overall likelihood is $L({\boldsymbol \theta} | ({\bf X,Y})^{\prime}) = \prod_{j,k} L( {\boldsymbol \theta}_{j,k}| \{ ( x,y)_{i,j,k} \}_{i} )$, where ${\boldsymbol \theta} = \{ {\boldsymbol \theta}_{j,k} \}_{(j,k)}$. 

We used independence priors for the mean and covariance parameters, $$ [ \{ (\mu_{x,j,k},\mu_{y,j,k},\sigma_{x,j,k},\sigma_{y,j,k},\rho_{j,k}) \}_{(j,k)}] = [ \{ (\mu_{x,j,k},\mu_{y,j,k} \} _{(j,k)}] [ \{\sigma_{x,j,k},\sigma_{y,j,k},\rho_{j,k}) \}_{(j,k)}] $$
where the components of each were modeled hierarchically. Specifically, for the mean parameters, we took $[ \{ \mu_{x,j,k},\mu_{y,j,k} \}  _{(j,k)} ] = [ \{ \mu_{x,j,k} \} _{(j,k)}] [ \{ \mu_{y,j,k} \}_{(j,k)}]$, which implies that the means for the $x$ and $y$ components of the vectors had independent prior distributions.  Across the $(j,k)$ combinations, the means for the components were assumed to be conditionally independent with $[ \{\mu_{x,j,k} \} | \mu_{x}, \sigma_{x}] \sim N(\mu_{x},\sigma^2_{x})$ and $[ \{ \mu_{y,j,k}\} | \mu_{y}, \sigma_{y}] \sim N(\mu_{y},\sigma^2_{y})$. We used fairly diffuse hyper-prior distributions $[\mu_x] = [\mu_y]\sim N(0,10)$ and $[\sigma^{-2}_x]=[\sigma^{-2}_y]\sim U(0,1)$.  Fixing the hyper-priors at specific constants resulted in no difference in the final results. 

The components of the covariance matrices were also modeled to have independent prior distributions, $$ [ \{\sigma_{x,j,k},\sigma_{y,j,k},\rho_{j,k}) \}_{(j,k)}] =[\{ \sigma_{x,j,k} \}_{(j,k)}][\{ \sigma_{y,j,k} \}_{(j,k)} ][\{ \rho_{j,k} \}_{(j,k)}]$$ with prior conditional independence for $(j,k)$ combinations and distributions  $[\sigma_{x,j,k} | a_x, b_x]\sim\Gamma(a_x,b_x)$, $[\sigma_{y,j,k}|a_y, b_y]\sim\Gamma(a_y,b_y)$ and $[a_x]=[a_y]\sim U(0,.5)$, $[b_x]=[b_y]\sim U(0,1)$, which give diffuse gamma distributions. Finally, the priors for the correlation parameters were also conditionally independent, $[\rho_{j,k}|\mu_{\rho},\sigma^2_{\rho}] \sim  N(\mu_{\rho},\sigma^2_{\rho})I_{\{-1,1\}}(\rho)$; that is, a truncated normal distribution with $[\mu_{\rho}]\sim N(0,2)$ and $[\sigma^{-2}_{\rho}] \sim U(0,1)$.  Fixing parameters of the priors to constants gave similar results to assigning hyper-prior distributions.   

The overall prior distribution for $\boldsymbol \theta$,  $ [{\boldsymbol \theta}] $, is given by a product of the prior and hyper-prior distributions, with the hyper-parameters $\boldsymbol{\eta} = (\mu_x, \mu_y, \sigma_x, \sigma_y, a_x, b_x, a_y, b_y, \mu_{\rho}, \sigma_{\rho})$ integrated out. The resulting posterior distribution for $\theta$ is obtained as 
$$ [ {\boldsymbol \theta} |  \bf{ (X,Y)}^{\prime}]= { {L({\boldsymbol \theta}|  \bf{ (X,Y)}^{\prime}) [{\boldsymbol \theta}]} \over { \int L({\boldsymbol \theta}|  \bf{ (X,Y)}^{\prime}) [{\boldsymbol \theta}]  d {\boldsymbol \theta}}} , $$
with the high dimensional integration in the denominator bypassed using Markov chain Monte Carlo (MCMC) sampling. 

Furthermore, we use the posterior distributions to obtain posterior predictive distributions for `new' observations from each of the 26 landmark-group combinations. Using predictive inference to assess the potential similarities and differences across the two groups at each landmark gives inference in terms of the observables, namely, the $\{(x,y)^{\prime}\}_{(j,k)}$ vectors instead of making inferences on the parameters in $\boldsymbol \theta$ used to model the data on the vectors. The predictive distributions are obtained as 
$$ [ (x,y)^{\prime}_{new,j,k} | {\bf (X,Y)}^{\prime}] = \int [  (x,y)_{new,j,k} | {\boldsymbol \theta} ] [{\boldsymbol \theta} |   {\bf (X,Y)}^{\prime}]   d {\boldsymbol \theta}$$
where  $[  (x,y)_{new,j,k} | {\boldsymbol \theta} ]$ is a bivariate normal distribution with 
 $ (x,y)_{new,j,k} $  conditionally independent of the observed data $\bf (X,Y)^{\prime}$. Again, MCMC is used to bypass this integration.

We used open source programs for all model implementation; specifically, R (R Development Core Team 2015) using the `rjags'$^{\textrm{TM}}$ (Plummer 2012b) package which interface with MCMC sampling programs `Just Another Gibbs Sampler (JAGS)'$^{\textrm{TM}}$ (Plummer 2003).  The `coda'$^{\textrm{TM}}$ package (Plummer 2012a) assists in managing and presenting sampling output. The `emdbook'$^{\textrm{TM}}$ package (Bolker 2014) was used to produce contour plots. 

\section{Acknowledgements}
The paper was completed while SH was in the Statistics Department at the University of Wyoming.


\end{document}